\documentclass[10pt,twocolumn,letterpaper]{article}

\usepackage[pagenumbers]{cvpr} 

\definecolor{cvprblue}{rgb}{0.21,0.49,0.74}
\usepackage[pagebackref,breaklinks,colorlinks,allcolors=cvprblue]{hyperref}
\usepackage{float}
\usepackage{hyperref}
\usepackage{url}
\usepackage{booktabs}
\usepackage{colortbl}
\usepackage{multirow}
\usepackage{graphicx}

\def\paperID{22891} 
\def\confName{CVPR}
\def\confYear{2026}

\title{CamPilot: Improving Camera Control in Video Diffusion Model with Efficient Camera Reward Feedback}

\def\authorBlock{
    Wenhang Ge\textsuperscript{$1$,$3$}\thanks{Equal contribution} \thanks{This work was conducted during the author's internship at Kling.}\qquad
    Guibao Shen\textsuperscript{$1$,$3$
    }\footnotemark[1] \footnotemark[2]  \qquad
    Jiawei Feng\textsuperscript{$1$}\footnotemark[1] \qquad
    Luozhou Wang\textsuperscript{$1$} \quad
    Hao Lu\textsuperscript{$1$} \quad \\
    Xingye Tian\textsuperscript{$3$} \quad
    Xin Tao\textsuperscript{$3$} \quad
    Ying-Cong Chen\textsuperscript{$1,2$} \thanks{Corresponding author}\quad
    
    \\
    \small$^1$ HKUST(GZ)\quad 
    \small$^2$ HKUST\quad 
    \small$^3$ Kling Team, Kuaishou Technology
    
}
\author{\authorBlock}

\begin{document}

\maketitle

\begin{figure*}
\centering
    \includegraphics[width=0.9\linewidth]{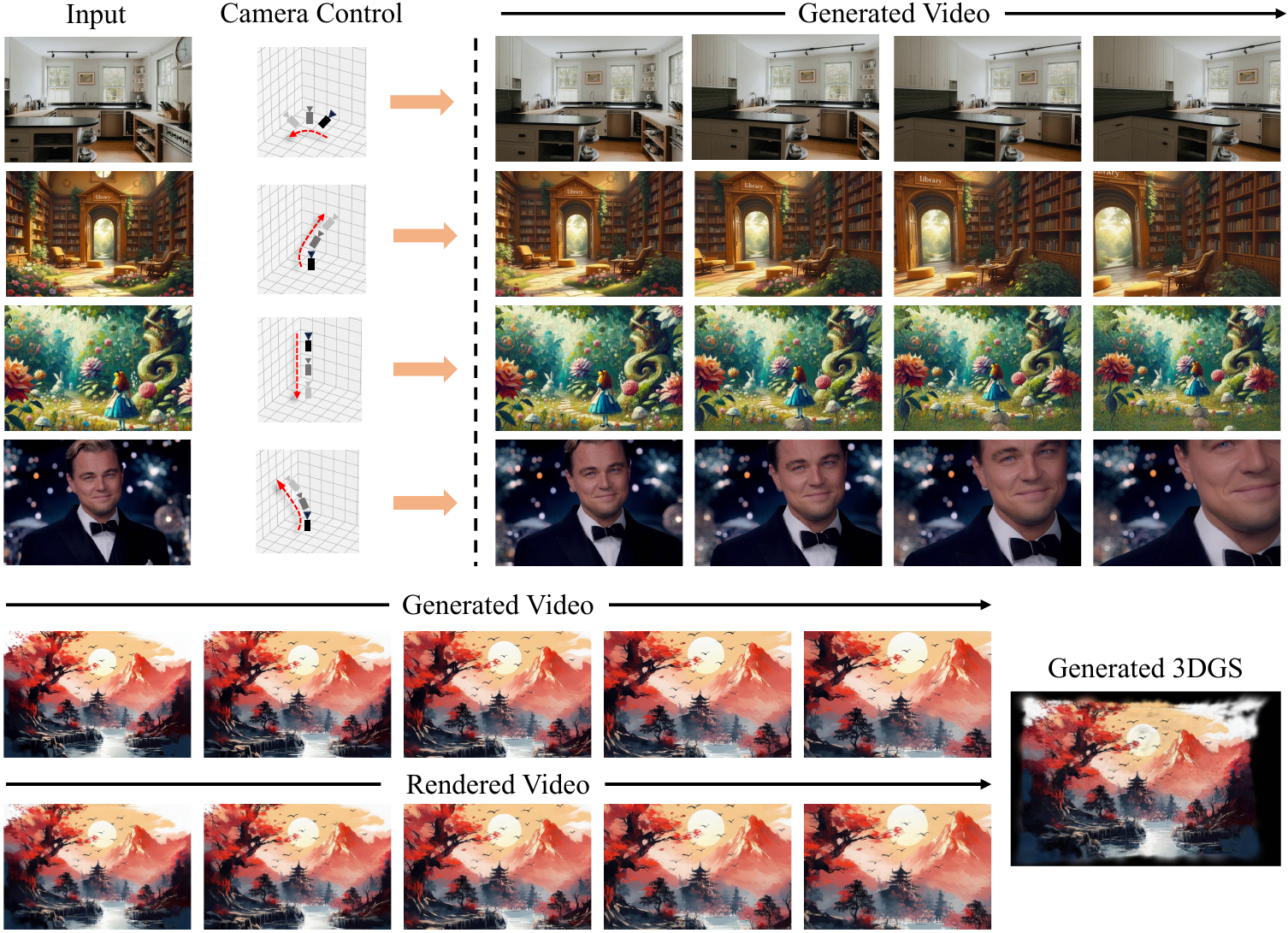}
    \vspace{-0.2cm}
    \captionof{figure}{Our model functions as a comprehensive framework for world-consistent video generation and scene reconstruction. In the upper section, it excels at generating 3D-consistent scene videos for world exploration by following custom camera trajectories. In the lower section, it efficiently reconstructs high-quality 3D scenes in a feed-forward manner with generated video frames.}
    \label{fig:teaser}
\end{figure*}

\begin{abstract}
Recent advances in camera-controlled video diffusion models have significantly improved video-camera alignment. However, the camera controllability still remains limited. In this work, we build upon Reward Feedback Learning and aim to further improve camera controllability. However, directly borrowing existing ReFL approaches faces several challenges. First, current reward models lack the capacity to assess video-camera alignment. Second, decoding latent into RGB videos for reward computation introduces substantial computational overhead. Third, 3D geometric information is typically neglected during video decoding. To address these limitations, we introduce an efficient camera-aware 3D decoder that decodes video latent into 3D representations for reward quantization. Specifically, video latent along with the camera pose are decoded into 3D Gaussians. In this process, the camera pose not only acts as input, but also serves as a projection parameter. Misalignment between the video latent and camera pose will cause geometric distortions in the  3D structure, resulting in blurry renderings. Based on this property, we explicitly optimize pixel-level consistency between the rendered novel views and ground-truth ones as reward. To accommodate the stochastic nature, we further introduce a visibility term that selectively supervises only deterministic regions derived via geometric warping. Extensive experiments conducted on RealEstate10K and WorldScore benchmarks demonstrate the effectiveness of our proposed method. Project page:
\href{https://a-bigbao.github.io/CamPilot/}{CamPilot Page}.

\end{abstract}    
\section{Introduction}
\label{sec:intro}

Video diffusion models have recently achieved impressive progress \cite{blattmann2023stable, yang2024cogvideox, liu2024sora}, enabling the generation of high-quality and temporally coherent videos conditioned on inputs such as text prompts or a single image. Despite these advances, real-world applications often demand a higher degree of controllability. A key factor is camera controllability. Users not only expect visually realistic content but also require explicit control over camera trajectories to support user-friendly and customizable content creation.

To address the need for camera-controlled video generation, several recent works \cite{yu2024viewcrafter, ren2025gen3c, gao2024cat3d, sun2024dimensionx, voleti2024sv3d, chan2023generative, sargent2024zeronvs, bahmani2024ac3d, he2024cameractrl, li2025realcam, zheng2024cami2v, yang2025omnicam} have explored this task by fine-tuning pretrained video models with camera conditioning. 
Recognizing that many downstream applications such as virtual reality \cite{schuemie2001research}, robotics \cite{mateo2016visual}, and game development \cite{gregory2018game} require not only high-quality visuals but also 3D/4D representations, these methods have begun to bridge the gap between 2D generation and 3D/4D reconstruction. 
Among them, static 3D scene generation is particularly valuable for architectural visualization, virtual tours, and digital content creation, where scene geometry remains consistent as demonstrated by WorldLabs \cite{li2025worldlabs}. 
A common strategy is to optimize over generated novel views.
Despite these advancements,  precise camera control is still difficult to achieve, often resulting in suboptimal convergence. 

Recent works \cite{prabhudesai2024video, li2024controlnet++, liu2025improving, zhang2024unifl, xu2023imagereward, prabhudesai2023aligning} have introduced Reward Feedback Learning (ReFL) for diffusion models to further refine the model according to  task-specific objectives, drawing inspiration from the Reinforcement Learning from Human Feedback (RLHF) \cite{grattafiori2024llama, yang2024qwen2, lee2023rlaif} of large language models (LLMs). 
For instance, VADER~\cite{prabhudesai2024video} explores a range of reward functions—such as perceptual quality, text-video semantic alignment, and aesthetic appeal—to enhance visual fidelity and semantic consistency. Controlnet++ \cite{li2024controlnet++} leverages pixel-level cycle consistency as a reward to improve image-based controllability. However, none of these approaches considers camera controllability.

In this work, we aim to enhance the adherence to camera conditioning through ReFL, a topic that remains under-explored in the context of video diffusion. However, there are three main challenges in adopting this strategy for camera-controlled video diffusion. First, current models struggle to assess the alignment of camera conditions in video generation. Second, reward computation necessitates decoding the generated latent into video, leading to VRAM inefficiency due to the resource-intensive nature of video decoders. Lastly, these methods often overlook the underlying 3D geometric structure during video decoding, which restricts their effectiveness. The task essentially requires the model to understand 3D world.

A naive approach would be to use COLMAP \cite{schoenberger2016sfm} for camera pose estimation. However, the heavy computational cost and scale-invariant estimation make it infeasible for efficient training and precise pose supervision.
Considering the three challenges, we introduce a camera-aware 3D decoder that enables computationally efficient evaluation of video-camera consistency without requiring heavy computation. 
Specifically, we project the video latent—obtained by encoding a raw video using the video VAE—along with the corresponding ground-truth camera poses into a 3D representation, namely 3D Gaussians (3DGS) \cite{kerbl3Dgaussians}. This representation supports efficient novel view rendering from arbitrary viewpoints and utilizes photometric loss for supervision.
In this projection process, camera poses play a crucial role. On the one hand, they are transformed into Plücker embeddings \cite{he2024cameractrl} as part of the network input. On the other hand, the mean of each 3D Gaussian is computed by projecting the camera pose along with the predicted depth.
These two mechanisms ensure that when the generated video latent is not aligned with the camera condition, the resulting 3DGS becomes geometrically inconsistent, leading to degraded renderings.
Based on this property, we regard minimizing the pixel-level difference between the rendered videos and ground-truth sequences as reward. This design is consistent with the nature of the proposed camera-aware 3D decoder, which emphasizes low-level visual cues.

However, computing pixel-level rewards presents unique challenges. High-level semantic rewards can be meaningfully applied across multiple diverse diffusion samples, while low-level pixel alignment rewards are sensitive to diverse generation results. 
Camera-controlled video generation often involves hallucinated content, making it difficult to enforce strict pixel-level consistency across all pixels without suppressing generative diversity. 
To address this, our reward formulation is carefully designed to focus only on deterministic regions that are visible in the conditioning image, while ignoring unconstrained areas that permit creative generation.
To this end, we design a visibility-aware reward objective that restricts reward computation to deterministic regions while avoiding penalization in hallucinated or occluded areas. Since our camera-aware 3D decoder is inherently 3D-aware, we can render depth maps from the 3DGS. By combining the rendered depth with camera poses, we can determine the visibility of each pixel across all frames through geometric warping.

To summarize, our contributions are listed as follows.
\begin{itemize}
    \item We leverage an efficient camera-aware 3D decoder for reward computation. This module lifts the video latent and the camera pose into 3DGS, which supports rendering to 2D images and enables the evaluation of the alignment between camera conditions and the generated video.
    \item We propose camera reward optimization to improve the alignment between the generated video and camera conditions by minimizing the deterministic pixel-level differences between the rendered and ground-truth videos.
    \item Extensive experiments demonstrate the effectiveness of the proposed framework, significantly improving camera controllability and visual quality.
\end{itemize}

\section{Related Work}
\label{sec:formatting}

\label{related work}

\subsection{Camera controlled Video Diffusion Models}
With the rapid advancements in video diffusion models, camera-controlled video generation \cite{he2024cameractrl, bahmani2024vd3d, bahmani2024ac3d, yu2024viewcrafter, ren2025gen3c, gao2024cat3d, wang2024motionctrl, hu2025ex, li2025realcam, zheng2024cami2v, yang2025omnicam, bai2025recammaster} has garnered significant attention in the research community. 
Recent works such as MotionCtrl \cite{wang2024motionctrl}, CameraCtrl \cite{he2024cameractrl}, and ViewCrafter \cite{yu2024viewcrafter} inject various forms of camera conditioning—ranging from extrinsics and Plücker embeddings \cite{sitzmann2021light} to point cloud renders—into pretrained video generation models. More recently, AC3D \cite{bahmani2024ac3d} has carefully explored the spatial and temporal points at which camera representations should be injected. 
CamCo \cite{xu2024camco} introduces epipolar constraints into attention layers, while  Gen3C \cite{ren2025gen3c} and FlexWorld \cite{chen2025flexworld} maintain a spatiotemporal 3D cache to enhance robustness in camera control.
Besides static scene generation, several works \cite{he2025cameractrl, bai2025recammaster, wu2025cat4d} also explore camera control in dynamic scenes. 
For example, CameraCtrl2 \cite{he2025cameractrl} investigates this task from a dataset curation perspective to enable dynamic scene generation with controllable cameras. 
These methods address different scenarios and challenges compared to our focus on static scene generation.
Despite these advances, existing approaches still face challenges in achieving precise control. In this work, we enhance camera controllability by reward feedback learning. 
Note that though our framework adopts Plücker embeddings as camera condition, the proposed reward feedback learning method is a general method and can be applied to any form of camera condition representation.

\subsection{3D Generative Models}
Object-level 3D generative models \cite{hong2023lrm, GS-LRM, ge2024prm, ge2023ref, zhang2024clay, jiang2025dimer, xu2024flexgen} have made remarkable progress in recent years, largely driven by the availability of large-scale 3D object datasets. 
However, 3D scene generation remains relatively under-explored. 
Most video diffusion based approaches \cite{yu2024viewcrafter, ren2025gen3c, gao2024cat3d, sun2024dimensionx, voleti2024sv3d, chan2023generative, sargent2024zeronvs} typically adopt a two-stage pipeline for 3D scene generation. In the first stage, diffusion models are employed to generate novel views given sparse or single-view observations and target poses. In the second stage, per-scene optimization is conducted using the generated novel views and corresponding target poses. 
For example, ViewCrafter \cite{yu2024viewcrafter} leverages a pretrained 3D foundation model to estimate point clouds, and subsequently employs a video diffusion model to perform inpainting with partial point cloud renderings.
Despite their effectiveness, such two-stage approaches suffer from two main limitations. First, the per-scene optimization process is time-consuming, making it difficult to scale to large numbers of scenes. Second, the quality of scene reconstruction is highly sensitive to the consistency between the generated novel views and the target camera poses. Misalignment between them can lead to suboptimal convergence. 
In contrast, we leverage a camera-aware 3D decoder that not only enables efficient 3D scene reconstruction in a feed-forward manner, but also serves as a media for reward computation by minimizing the misalignment between generated novel views and their corresponding target poses.

\subsection{Aligning Diffusion Models with Preference} 
Drawing inspiration from Reinforcement Learning from Human Feedback (RLHF) in the field of large language models (LLMs), recent works have begun to incorporate similar paradigms into diffusion models to better align generation quality with human preferences \cite{yang2024using, prabhudesai2024video, yuan2024instructvideo, liu2025improving, li2024t2v, li2024controlnet++, xu2023imagereward, zhang2024unifl}. For instance, ControlNet++ \cite{li2024controlnet++} explicitly optimizes pixel-level cycle consistency between generated images and conditional controls for improving controllable generation. UniFL \cite{zhang2024unifl} proposes a unified framework that
leverages feedback learning to enhance diffusion models comprehensively.
VADER \cite{prabhudesai2024video} explores a variety of reward models to fine-tune video generation.
However, these approaches require decoding the video latent into RGB video as input for the reward model to compute the reward gradient. This process introduces significant memory costs, constraining efficiency. Moreover, while these methods primarily focus on enhancing overall quality or alignment with text prompts, none explicitly address the challenge of improving camera controllability in video generation.
To address this gap, we leverage a novel camera-aware 3D decoder specifically designed to enhance camera controllability in video diffusion models through reward feedback learning.
\begin{figure*}[h]
    \centering
    \includegraphics[width=0.98\linewidth]{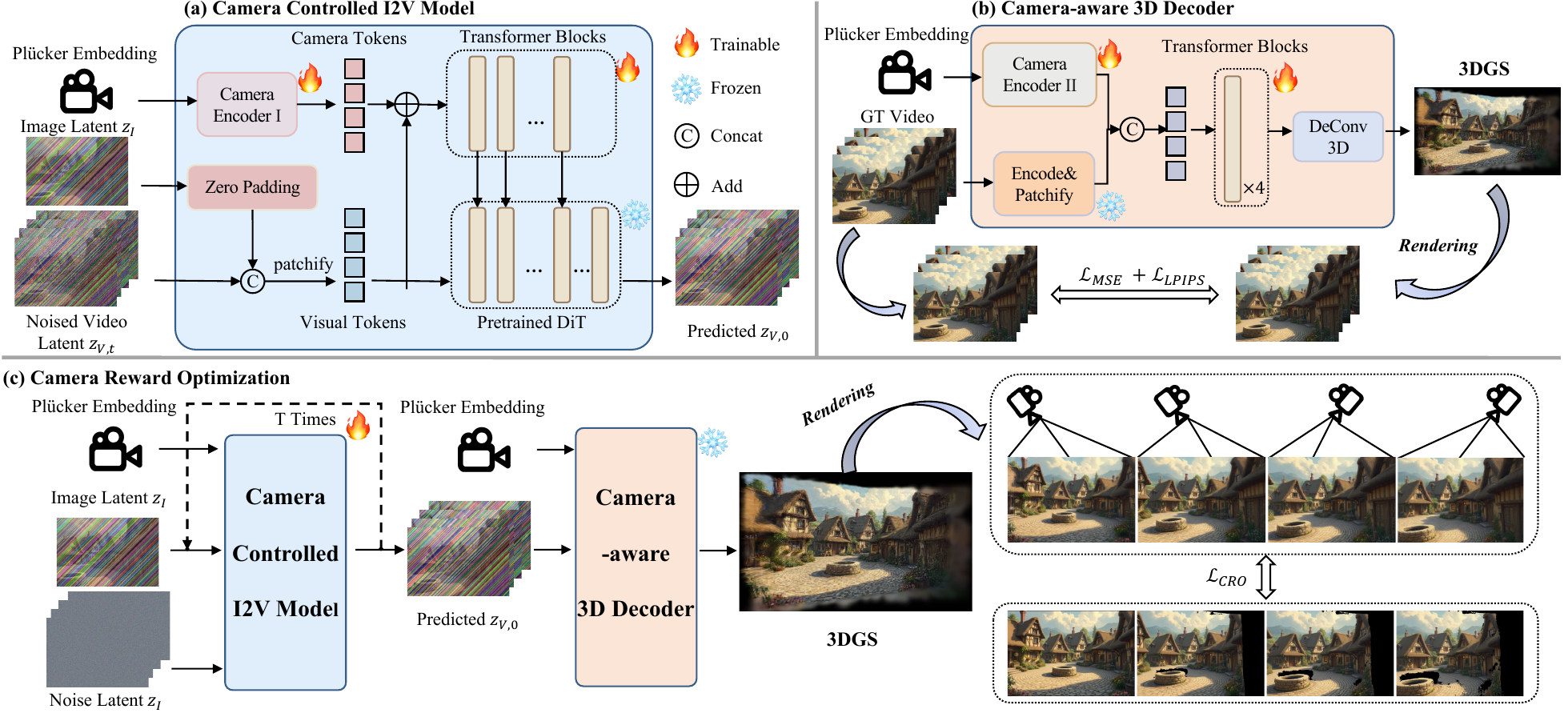}
    \vspace{-0.3cm}
    \caption{Overall of our framework. It consists of (a): a camera-controlled I2V model, where we inject Plücker Embedding as camera condition using ControlNet. (b) A camera-aware 3D decoder that decodes latent to 3DGS, supporting rendering for reward computation. (c) Camera reward optimization that minimizes mask-aware difference between rendered videos and ground-truth ones.}
    \label{fig:overall}
\end{figure*}
\section{Method}

We begin with a brief overview of camera-controlled video diffusion models and reward feedback learning in Section~\ref{Preliminaries}. Section~\ref{sec2} describes the training of the camera-controlled video diffusion model. We then introduce camera-aware 3D decoder in Section~\ref{sec3}, followed by the camera reward optimization in Section~\ref{sec4}. An overview of the entire framework is shown in Figure~\ref{fig:overall}.

\subsection{Preliminaries}
\label{Preliminaries}

\noindent\textbf{Camera controlled video diffusion model} learns to model the conditional distribution $p(\mathbf{x}_0|c, \mathbf{s})$, where $\mathbf{x}_0$ denotes the video latent  obtained from a video VAE \cite{yang2024cogvideox} 
, $c$ refers to the text or image condition and $\mathbf{s}$ is the camera condition. During training, noise $\epsilon_t$ is added to the latent $\mathbf{x}_0 $ at each timestep $t \in [0, T]$ and  a transformer model \cite{Peebles2022DiT} is optimized to predict this noise using the following
objective:
\begin{equation}
L(\theta) = \mathbb{E}_{\mathbf{x}_0, \epsilon, c, \mathbf{s}, t} \left[ \| \epsilon - \hat{\epsilon}_\theta(\mathbf{x}_t, c, \mathbf{s}, t) \|_2^2 \right].
\end{equation}
Following prior methods \cite{he2024cameractrl, bahmani2024ac3d, bahmani2024vd3d}, we adopt the Plücker embedding \cite{sitzmann2021light} as the camera condition, which provides pixel-aligned camera information and facilitates the use of ControlNet \cite{zhang2023adding} for conditioning.  

\vspace{0.1cm}
\noindent\textbf{Reward feedback learning} is a preference fine-tuning framework that directly optimizes the generation process using differentiable reward models and aims to improve the  model by aligning the behavior of network output with external preference signals, such as human feedback or heuristic reward models \cite{wallace2024diffusion, black2023training, xu2023imagereward}. 

\subsection{Adding Camera Control to Video Generation}
\label{sec2}
Following previous works \cite{he2024cameractrl, bahmani2024vd3d, bahmani2024ac3d, liang2024wonderland}, we incorporate camera information (i.e., Plücker embeddings) into the denoising process through ControlNet \cite{zhang2023adding} as shown in Fig. \ref{fig:overall} (a).
The Plücker embeddings are first compressed along the spatial and temporal dimensions to align with the shape of the video latent. To construct the ControlNet, we replicate the transformer blocks from the base video model and append a zero-initialized linear layer for stable training. 
Inspired by AC3D \cite{bahmani2024ac3d}, we copy only the first several transformer blocks, which has been shown to strike a balance between controllability and computational efficiency.
AC3D further observes that video diffusion models tend to establish low-frequency camera motion during the early timestep of the denoising process. Injecting camera condition at later timesteps provides limited benefits and may even impair visual quality. Following this insight, we adopt a truncated normal distribution with a mean of 0.8 and a standard deviation of 0.075, restricted to the interval $[0.6, 1]$, to bias timestep sampling toward earlier denoising steps where camera control is most effective. The network architecture and training details can be found in the Supplementary.

Despite these advancements, the overall camera controllability remains limited. 
In this work, we aim to further improve the alignment by explicitly optimizing over denoising trajectories with reward feedback learning.
To enable this, we first introduce an efficient camera-aware 3D decoder that quantitatively evaluates the alignment of camera trajectory in the generated videos. This is followed by a dedicated camera reward optimization.

\subsection{Camera-aware 3D decoder}
\label{sec3}
ReFL methods typically require decoding the latent into RGB frames for reward computation.
However, this process is computationally expensive and captures only 2D information, whereas camera-controlled video generation inherently requires the model to reason about 3D geometry.
Therefore, we aim to develop an efficient 3D decoder that reconstructs videos from latent while explicitly incorporating 3D information.
Moreover, the decoder should be camera-aware, enabling quantitative evaluation and optimization of video–camera alignment.

To meet these requirements, we leverage a latent-based feed-forward 3D Gaussian model as our camera-aware 3D decoder. 
Specifically, we train a transformer that takes both video latent and their corresponding Plücker embeddings as input, and outputs per-pixel aligned 3DGS. 
The positions of these 3DGS are estimated by projecting the camera parameters together with the predicted ray distances \( t \) using the relation \( \mathbf{u} = \mathbf{r_o} + t \cdot \mathbf{r_d} \). 
To train the decoder, we randomly select a stride \( s \) to sample a video sequence consisting of \( T \) frames. The video VAE encoder first compresses these \( T \) frames into latent, which are subsequently fed into transformer blocks along with Plücker embeddings to predict the 3DGS. 
The training objective employs a combination of mean squared error (MSE) loss and LPIPS loss~\cite{lipis} as shown in Fig. \ref{fig:overall} (b). 
The detailed architecture and training strategy can be found in the Supplementary. 

\begin{figure}
    \centering
    \includegraphics[width=1.\linewidth]{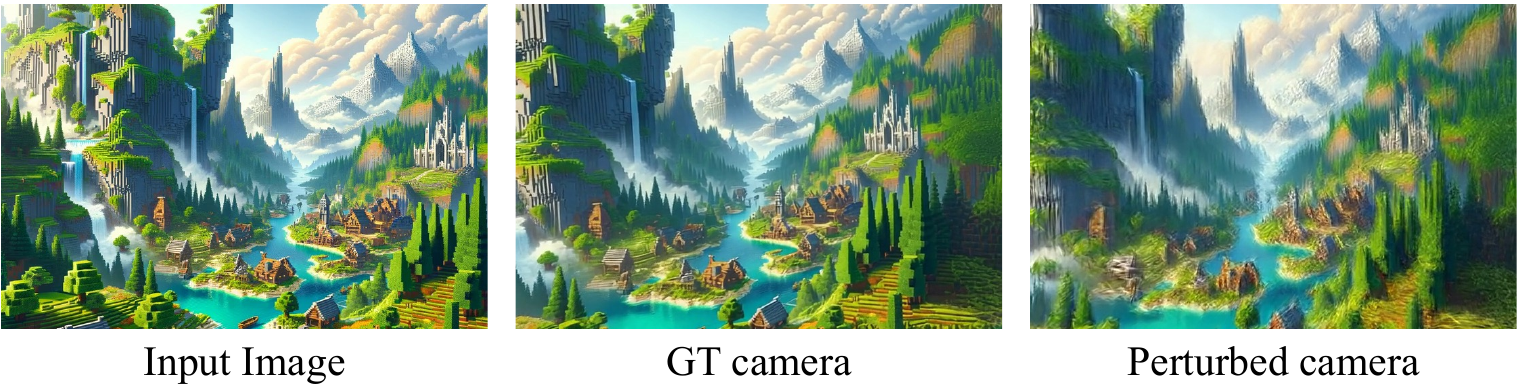}
    \vspace{-0.4cm}
    \caption{We add perturbation to the GT camera pose, and the rendered image becomes noticeably blurred, indicating the importance of aligned poses for rendering photorealistic images. }
    \label{camera perb}
\end{figure}

Within this framework, the camera poses play a crucial role, which act as input and projection variable. 
As a result, if the input latent and the camera poses are not well aligned, the 3D geometry deteriorates, leading to noticeably blurrier rendering. We show an example in Fig. \ref{camera perb}. Based on this property, we design our reward for feedback learning.

Note that our camera-aware 3D decoder is conceptually similar to Wonderland \cite{liang2024wonderland}, which designs a latent large reconstruction model (LaLRM) for 3D reconstruction from video latent.
However, our objective is fundamentally different. We employ the camera-aware 3D decoder for camera reward optimization, aiming to further enhance video–camera alignment, whereas Wonderland uses LaLRM solely for 3D reconstruction.

\subsection{Camera Reward Optimization}
\label{sec4}
With the camera-aware 3D decoder, we propose Camera Reward Optimization (CRO) to use reward gradients to further improve the camera controllability. 
Considering the property that if the generated videos misalign with the camera condition, the renderings become blurry, a naive approach is to penalize the blurriness. However, directly penalizing the blurriness easily leads to reward hacking issues \cite{skalse2022defining}. 
Hence, we regard minimizing the pixel-level difference between the rendered videos and ground-truth sequences as the reward. 

However, video generation introduces inherent stochasticity, making it infeasible to directly minimize the pixel-level difference since newly generated parts cannot align with ground-truth videos. 
To accommodate the stochastic nature, we adopt a visibility-aware reward strategy that restricts supervision to pixels that are visible in the conditioning image, which is generally deterministic. Visible mask can be derived by geometric warping, which requires depth information and camera poses. Fortunately, due to the inherent 3D structure of our camera-aware 3D decoder, we can obtain the rendered depth, which facilitates visibility estimation.
Specifically, given the ground-truth video with corresponding camera poses \( \mathbf{E} = [R; t] \in \mathbb{R}^{T \times 3 \times 4} \) and intrinsic matrix \( K \in \mathbb{R}^{T \times 3 \times 3} \), and the image condition \( \mathbf{I}_0 \in \mathbb{R}^{H \times W \times 3} \), which is the first frame of the video, we compute a per-frame visibility mask based on geometric warping. Omitting the temporal script, each pixel \( (u, v) \) from the target view is back-projected into 3D world coordinates using the rendered depth map \( \mathbf{D} \), intrinsic matrix \( \mathbf{K} \), and camera extrinsic matrix $\mathbf{E}$:
\begin{equation}
\mathbf{X}^{\text{world}}(u, v) = \mathbf{E} \cdot 
\begin{bmatrix}
\mathbf{D}(u, v) \cdot \mathbf{K}^{-1} [u, v]^T \\
\end{bmatrix}.
\end{equation}
Next, the 3D points are projected into the conditioned reference view using the its extrinsic matrix $\mathbf{E}_0$ and intrinsic matrix $\mathbf{K}_0$. The projected 2D coordinates in the reference view are obtained by:
\begin{equation}
\mathbf{x}^{(0)}(u, v) = \mathbf{K}_0 \cdot \mathbf{E}_0^{-1} \cdot 
\begin{bmatrix}
\mathbf{X}^{\text{world}}(u, v) \\
\end{bmatrix}.
\end{equation}
We then sample the reference depth map $\mathbf{D}_0$ at the projected location to obtain ${D}_0^{\text{proj}}(u, v)$. A visibility mask $\mathbf{M}$ is constructed by comparing the reprojected depth $\hat{z}^{(0)}(u, v)$ with the sampled depth, and a pixel is considered visible if the two depths agree within a tolerance $\tau$:
\begin{equation}
 {M}(u, v) = 
\begin{cases}
1, & \text{if } \left| \hat{z}^{(0)}(u, v) - {D}_0^{\text{proj}}(u, v) \right| < \tau \\
0, & \text{otherwise}.
\end{cases}
\end{equation}
With the visibility mask, we follow the VADER framework \cite{prabhudesai2024video} and restrict the reward on deterministic pixels, defining a masked MSE loss and LPIPS loss between the rendered image $\hat{\mathbf{I}}$ and the ground-truth image $\mathbf{I}$ as:
\begin{equation}
    \mathcal{L}_{\text{CRO}} = \mathcal{L}_{\text{MSE}}(\hat{\textbf{I}}, \textbf{I}, \textbf{M}) + \lambda \cdot \mathcal{L}_{\text{LPIPS}}(\hat{\textbf{I}}, \textbf{I}, \textbf{M}),
\end{equation}
where $\lambda$ is set empirically to 0.5.




\begin{figure*}[ht]
    \centering
    \includegraphics[width=0.9\linewidth]{ 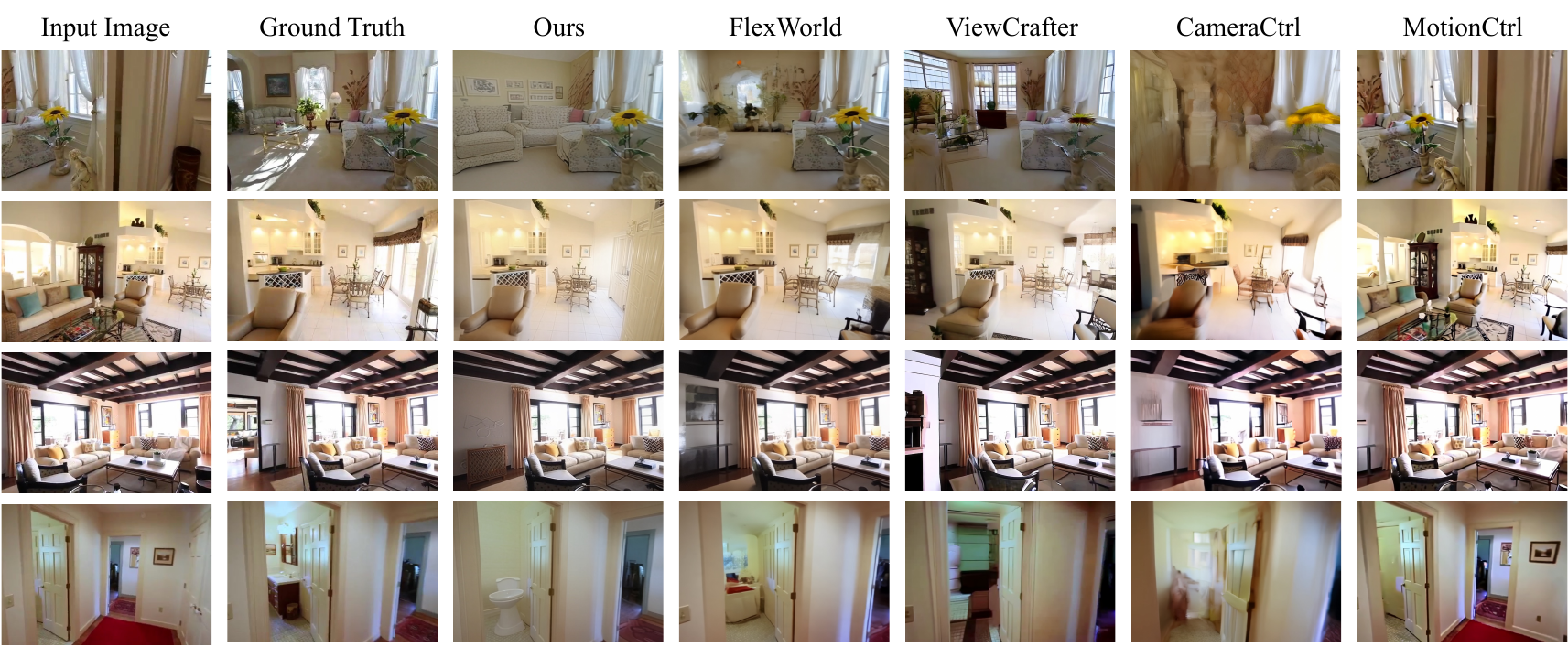}
    \vspace{-0.3cm}
    \caption{Qualitative comparison of video generation: our model produces novel views that are better aligned with the camera poses. }
    \label{fig:video_comparison}
\end{figure*} 

\begin{table*}[h]
\centering
\footnotesize
\caption{Quantitative comparison on video and 3D scene generation with the baseline methods.}
\vspace{-0.3cm}
\begin{tabular}{lccccccc|ccc}
\toprule
& \multicolumn{7}{c|}{\textbf{Video Generation}} & \multicolumn{3}{c}{\textbf{3D Scene Generation}} \\
\cmidrule(lr){2-8} \cmidrule(lr){9-11}
\textbf{Method} & \textbf{FID} $\downarrow$ & \textbf{FVD} $\downarrow$ & \textbf{$R_{\text{err}}$} $\downarrow$ & \textbf{$T_{\text{err}}$} $\downarrow$  & \textbf{PSNR} $\uparrow$ & \textbf{LPIPS} $\downarrow$ & \textbf{SSIM} $\uparrow$ & \textbf{PSNR } $\uparrow$ & \textbf{LPIPS} $\downarrow$ & \textbf{SSIM } $\uparrow$ \\
\midrule
Rec-only & - & - & - & - & - & - & - & 27.57 & 0.181 & 0.883 \\
\midrule
MotionCtrl  & 24.67 & 205.27 & 0.153 & 0.385 & 14.24 & 0.520 & 0.532 & 14.02 & 0.536 & 0.533 \\
CameraCtrl  & 22.17 & 96.52 & 0.078 & 0.222  & 17.58 & 0.586 & 0.360 & 17.30 & 0.391 & 0.573 \\
ViewCrafter  & 17.92 & 109.30 & 0.039 & 0.194 & 19.33 & 0.326 & 0.710 & 18.57 & 0.383 & 0.688 \\
FlexWorld  & 17.23 & 103.94 & 0.030 & 0.177 & 21.27 & 0.292 & 0.731 & 19.12 & 0.360 & 0.703 \\
\rowcolor{gray!20} Ours & \textbf{11.22} & \textbf{81.35} & \textbf{0.023} & \textbf{0.152} & \textbf{23.77} & \textbf{0.226} & \textbf{0.766} & \textbf{21.72} & \textbf{0.272} & \textbf{0.717} \\
\bottomrule
\end{tabular}

\label{tab:combined_comparison}
\end{table*}
\section{Experiments}

\subsection{Datasets and Evaluation Protocol}
\noindent\textbf{Training Datasets.}
Following previous methods \cite{bahmani2024vd3d, wang2024motionctrl, he2024cameractrl}, we utilized RealEstate10K (RE10K) \cite{zhou2018stereo} as our training data for fair comparison, which contains approximately 65K videos in the train split, primarily depicting static real estate environments. 
We used these 65K videos for training both the camera-aware 3D decoder and the camera-controlled video diffusion model.

\begin{table*}[]
  \footnotesize
  \centering
  \caption{Quantitative comparison across control and consistency metrics. Higher is better.}
  \vspace{-0.2cm}
  \label{tab:control_metrics}
  \begin{tabular}{llcccccccc}
    \toprule
    \multicolumn{2}{l}{\textbf{\shortstack{Methods}}} & \textbf{\shortstack{WorldScore \\Average}} & \textbf{\shortstack{Camera \\Control}} & \textbf{\shortstack{Object\\ Control}} & \textbf{\shortstack{Content\\ Alignment}} & \textbf{\shortstack{3D \\Consistency}} & \textbf{\shortstack{Photometric\\ Consistency}} & \textbf{\shortstack{Style \\Consistency}} & \textbf{\shortstack{Subjective \\Quality}} \\
    \midrule
    MotionCtrl  &  & 64.15 & 58.65 & 44.54 & 48.42 &  89.87 & 88.13 & 67.37 & 52.07 \\
    CameraCtrl &  & 65.42 & 65.72 & 45.31 & 49.10 & 90.07 & 92.42 & 64.70 & 50.64 \\
    ViewCrafter &  & 65.47 & 72.40 & 50.71 & 52.34 & 60.56 & 88.30 & 78.29 & 55.68 \\
    FlexWorld   &  & 71.35 & 68.16 & 56.15 & 53.66 & 84.43 & 91.31 & 86.07 & 59.65 \\
    \rowcolor{gray!20} Ours        &  & \textbf{74.45} & \textbf{86.26} & 49.75 & 46.46 & \textbf{90.64} & \textbf{93.30} & \textbf{89.78} & \textbf{64.95} \\
    \bottomrule
  \end{tabular}
  
  \label{ws}
\end{table*}

\begin{figure*}[ht]
    \centering
    \includegraphics[width=0.95\linewidth]{ 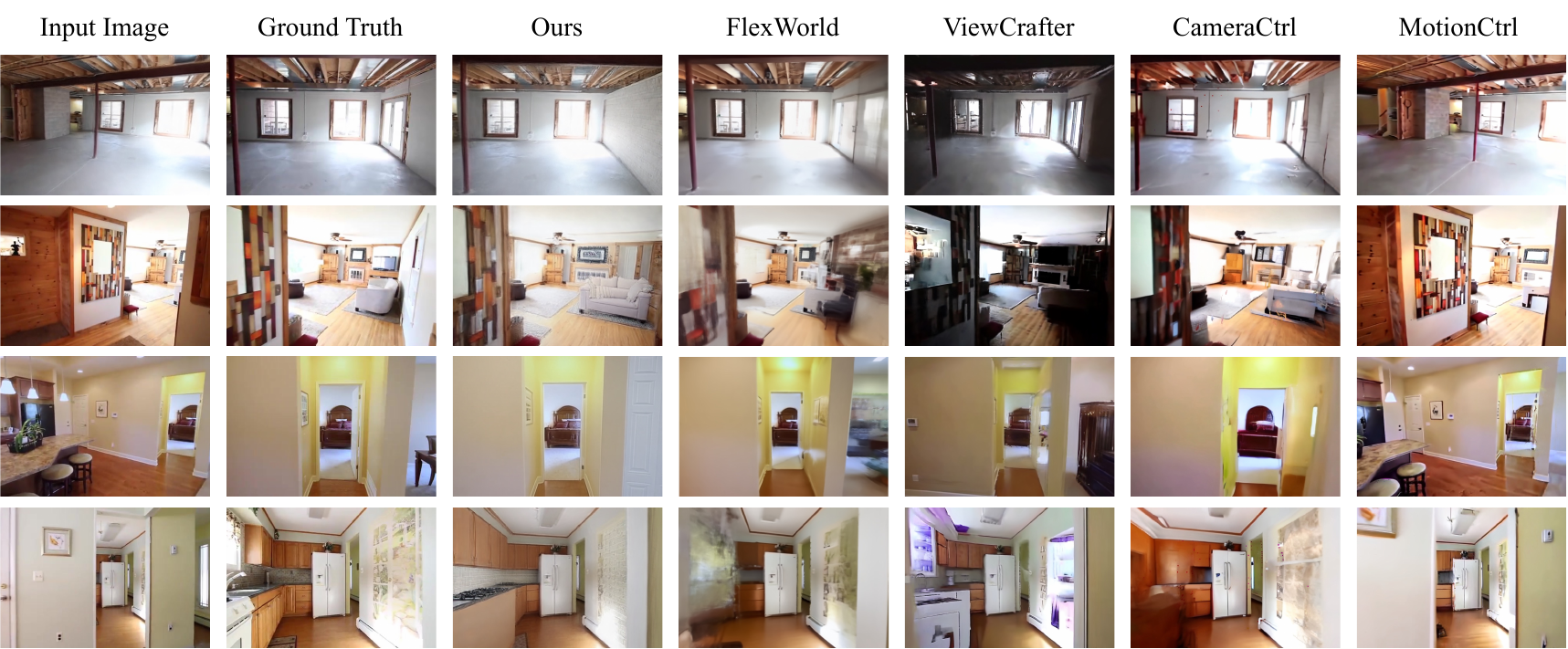}
    \vspace{-0.3cm}
    \caption{Qualitative comparison of 3D scene generation: our model produces more photorealistic novel view
    renderings that are aligned with the camera poses, outperforming other methods}
    \label{fig:compa_3d}
\end{figure*}

\noindent\textbf{Testing Datasets.}
Following previous works \cite{liang2024wonderland, yu2024viewcrafter}, we randomly selected 300 videos from the approximately 7K test sets of RE10K, ensuring no overlap with the training data.
We also adopted the WorldScore \cite{duan2025worldscore} static benchmark for out-of-domain comparison, which consists of 2,000 static test examples. In each example, an input image and a camera trajectory are provided.

\begin{figure*}[ht]
    \centering
    \includegraphics[width=0.91\linewidth]{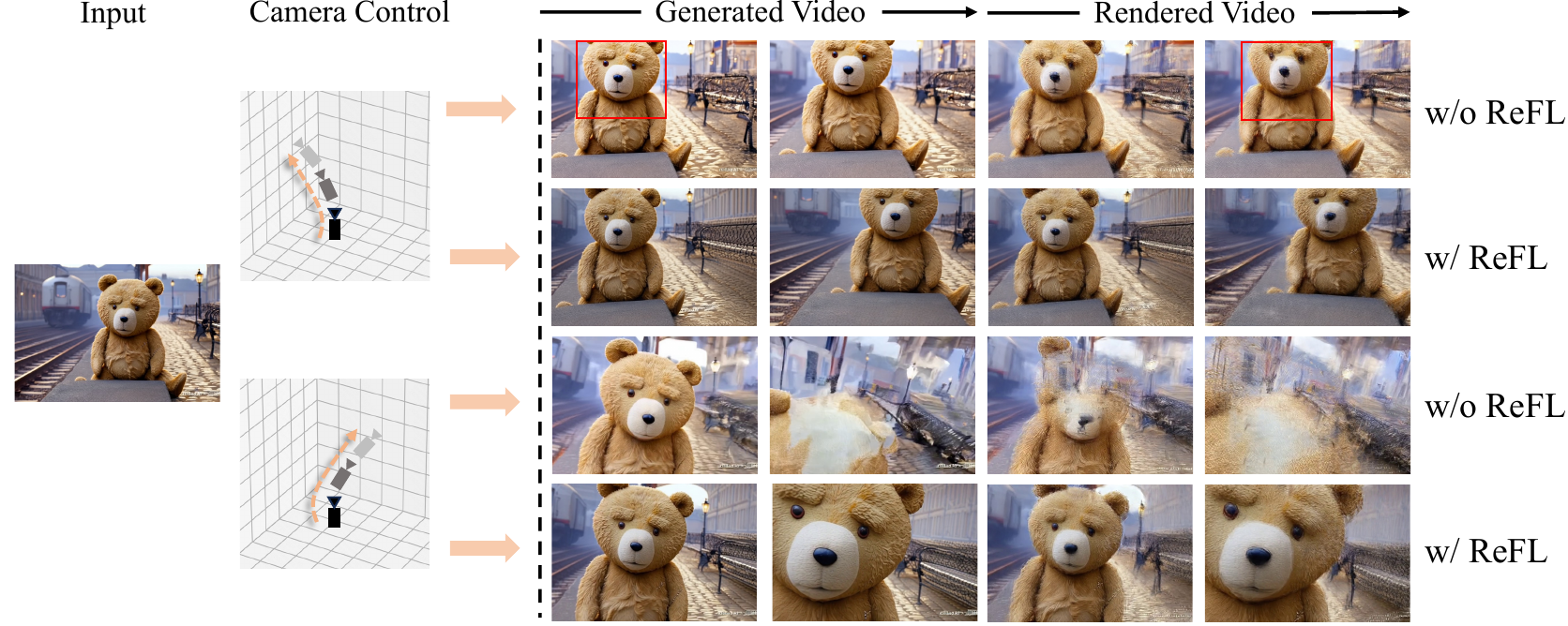}
    \vspace{-0.3cm}
    \caption{Ablation study to validate the effectiveness of ReFL. Before applying ReFL, the generated and rendered videos can exhibit severe artifacts, whereas the visual quality improves significantly after applying ReFL.}
    \label{fig:ablation}
\end{figure*}

\noindent\textbf{Evaluation Protocol.}
We evaluated the quality of the generated videos using multiple metrics. Following previous works \cite{liang2024wonderland, he2024cameractrl, bahmani2024vd3d, bahmani2024ac3d}, we employed Fréchet Inception Distance (FID) \cite{heusel2017gans} and Fréchet Video Distance (FVD) \cite{unterthiner2019fvd} to assess visual quality. Additionally, PSNR, LPIPS, and SSIM metrics were used to evaluate the quality of novel view synthesis, camera controllability, and the performance of scene reconstruction.
Following the approach in Wonderland \cite{liang2024wonderland}, we also compute these metrics for the first 14 frames due to the randomness in generation.
For further evaluating camera controllability, we used rotation error ($R_\text{err}$) and translation error ($T_\text{err}$) computed via DROID-SLAM \cite{teed2021droid} following WorldScore \cite{duan2025worldscore}. 
Furthermore, we evaluated WorldScore \cite{duan2025worldscore} on WorldScore static benchmark.
In addition, 
we compared the decoded video (from the video VAE decoder) and the rendered video (from the camera-aware 3D decoder) using the same generated latent. We report PSNR, SSIM \cite{wang2004image}, and LPIPS \cite{zhang2018unreasonable} as metrics to further evaluate the camera controllability.

\subsection{Implementation Details.}
We built our model upon CogVideoX-5B-I2V \cite{yang2024cogvideox}.
To inject camera conditioning, we adopt a ControlNet \cite{zhang2023adding}  with 8 blocks, initialized from the pretrained video model, followed by a zero-initialized linear projection layer.
For the camera-aware 3D decoder, we employ 4 transformer blocks with a hidden dimension of 1,024.
Please refer to the Supplementary for more implementation details.

\subsection{Comparison on Video Generation}
We compared the proposed framework with four baselines: MotionCtrl \cite{wang2024motionctrl}, CameraCtrl \cite{he2024cameractrl}, ViewCrafter \cite{yu2024viewcrafter}, and FlexWorld \cite{chen2025flexworld}. It is worth noting that our method focuses on static scene generation, consistent with all compared methods.
The first two methods directly use camera pose as conditioning, similar to our method. In contrast, the last two methods use point cloud renders as conditioning, which requires re-annotation of the camera pose to align the scale between the estimated point cloud and the camera pose. 
The qualitative comparison is illustrated in Fig. \ref{fig:video_comparison}, while the quantitative results are presented in Table \ref{tab:combined_comparison}. Our method surpasses existing approaches in both novel view synthesis and camera controllability.

\subsection{Comparison on Scene Generation}
To evaluate the effectiveness of our method for 3D scene generation, we compared the visual quality of the rendering results with the same four baseline methods using PSNR, LPIPS and SSIM between the renderings and ground-truth videos. 
We followed Wonderland \cite{liang2024wonderland} and only computed the metric in the first 14 frames subsequent to the conditional image. 
To evaluate the upper bound of our camera-aware 3D decoder, we also reported the PSNR, LPIPS and SSIM between ground-truth video and rendered video (denoted as ``Rec-only'') using video and ground-truth camera pose as input. The quantitative results are reported in Table \ref{tab:combined_comparison} and the qualitative comparison is illustrated in Fig. \ref{fig:compa_3d}.


\subsection{Comparison on WorldScore Benchmark}
We also compared on the WorldScore static benchmarks \cite{duan2025worldscore}. The quantitative results are reported in Table \ref{ws}. Additional qualitative comparisons are in the Supplementary. We reproduced the officially released code on this benchmark using the same test settings and hyperparameters.

\begin{table*}[h]
    \footnotesize
    \centering
    
    \caption{Ablation study to validate the effectiveness of each component.}
    \vspace{-0.3cm}
    \begin{tabular}{lccc cccc cccc}
        \toprule
       {Setting} & \multicolumn{3}{c}{Generated vs. GT} & \multicolumn{3}{c}{Rendered vs. GT} & \multicolumn{3}{c}{Rendered vs. Generated} \\
         & \multicolumn{3}{c}{\hrulefill} & \multicolumn{3}{c}{\hrulefill} & \multicolumn{3}{c}{\hrulefill} \\
        {Metric} & PSNR$\uparrow$ & LPIPS$\downarrow$ & SSIM$\uparrow$ & PSNR$\uparrow$ & LPIPS$\downarrow$ & SSIM$\uparrow$ & PSNR$\uparrow$ & LPIPS$\downarrow$ & SSIM$\uparrow$ \\
        \midrule
        w/o ReFL & 21.57 & 0.282 & 0.720 & 18.93 & 0.361 & 0.642 & 24.34 & 0.231 & 0.798 \\
        w/o visibility mask & 22.75 & 0.241 & 0.749 & 20.52 & 0.293 & 0.694 & 26.14 & 0.219 & 0.815 \\
        w/o novel view & 22.88 & 0.232 & 0.756 & 20.88 & 0.279 & 0.706 & 26.45 & 0.202 & 0.824 \\
        w/ CFG & 23.30 & 0.235 & 0.751 & 21.04 & 0.282 & 0.709 & 27.08 & 0.193 & 0.841 \\
        \rowcolor{gray!20} Full model &\textbf{23.77}  &\textbf{0.226}  &\textbf{0.766}  &\textbf{21.72}  &\textbf{0.272}  &\textbf{0.717}  &\textbf{27.13}  &\textbf{0.192}  &\textbf{0.844}  \\
        \bottomrule
    \end{tabular}
    \label{tab:ablation}
\end{table*}

\subsection{Ablation Study}
We conducted an ablation study to validate the effectiveness of each component in our framework. The quantitative results are presented in Table~\ref{tab:ablation}, using PSNR, SSIM, and LPIPS metrics. These metrics compare the decoded video (from the video VAE decoder) and the rendered video (produced by the reward model) from the same generated latent, denoted as ``Rendered vs Generated.'' Additionally, they compare the generated videos and rendered videos with the ground-truth ones, denoted as ``Generated vs GT'' and ``Rendered vs GT,'' respectively. 


\noindent\textbf{The effectiveness of reward feedback learning.}
Reward feedback learning (ReFL) is crucial for enhancing the camera controllability. We compared the results before and after applying ReFL (denoted as ``w/o ReFL'') in Table \ref{tab:ablation}. After implementing ReFL, the performance significantly improves, indicating that the reward gradients is effective and can further enhance camera controllability. We visualized a qualitative comparison in Fig. \ref{fig:ablation}. We discuss more insights in the Supplementary.

\noindent\textbf{The effectiveness of visibility mask.}
The visibility mask plays a crucial role in accommodating the stochastic nature of generative models by supervising only the deterministic pixels in the conditioning image. We conducted an experiment without using the visibility mask (denoted as ``w/o visibility mask'') as shown in Table \ref{tab:ablation}. The performance deteriorates without the visibility mask.

\noindent\textbf{The effectiveness of novel views.}
Unlike the video decoder, our proposed camera-aware 3D decoder can render novel views in addition to the seen views that are input to the video encoder. This capability allows us to incorporate novel views as supervision. We conducted an ablation study to validate the effectiveness of using novel views, denoted as ``w/o novel view'' in Table \ref{tab:ablation}. The performance of ``w/o novel view'' degrades, indicating the effectiveness of incorporating 3D geometric information.


\noindent\textbf{The effect of class free guidance.}
During each denoising step, we have the option to use class-free guidance (CFG) or not. We conducted an ablation study to assess the impact of CFG on sampling. The qualitative comparison is presented in Table \ref{tab:ablation}, labeled as ``w/ CFG''. The performance is comparable to that without CFG. However, since CFG results in twice the computational overhead during training, we have opted to disable CFG in our experiments.


\section{Conclusion and Limitation}
\noindent\textbf{Limitation.}
Despite the effectiveness, several limitations remain. First, the performance of the 3D decoder determines the upper bound of ReFL. For efficiency, we used only a 4 transformer blocks and trained solely on RE10K for fair comparison with previous methods. Scaling up the network and dataset may further improve this upper bound. 
Second, we only focus on static scene generation in this work. 3DGS can only represent static scenes and is not suitable for dynamic scene reconstruction. Exploring 4DGS as a reward model is a direction for future work. 

\vspace{0.1cm}
\noindent\textbf{Conclusion.}
In this work, we investigate the problem of camera-controlled video diffusion models and 3D scene generation, where the quality heavily relies on the alignment between camera conditions and the generated videos. 
To further improve this alignment, we introduce a camera-aware 3D decoder for efficient decoding video latent to rendered videos for reward computation.
During camera reward optimization, we propose to aligns the deterministic pixels between rendered videos and ground-truth videos to improve the rendering quality. Extensive experiments validate the effectiveness of the proposed method, outperforming existing methods by a large margin.

{
    \small
    \bibliographystyle{ieeenat_fullname}
    \bibliography{main}

\begin{thebibliography}{66}
\providecommand{\natexlab}[1]{#1}
\providecommand{\url}[1]{\texttt{#1}}
\expandafter\ifx\csname urlstyle\endcsname\relax
  \providecommand{\doi}[1]{doi: #1}\else
  \providecommand{\doi}{doi: \begingroup \urlstyle{rm}\Url}\fi

\bibitem[Bahmani et~al.(2024{\natexlab{a}})Bahmani, Skorokhodov, Qian, Siarohin, Menapace, Tagliasacchi, Lindell, and Tulyakov]{bahmani2024ac3d}
Sherwin Bahmani, Ivan Skorokhodov, Guocheng Qian, Aliaksandr Siarohin, Willi Menapace, Andrea Tagliasacchi, David~B Lindell, and Sergey Tulyakov.
\newblock Ac3d: Analyzing and improving 3d camera control in video diffusion transformers.
\newblock \emph{arXiv preprint arXiv:2411.18673}, 2024{\natexlab{a}}.

\bibitem[Bahmani et~al.(2024{\natexlab{b}})Bahmani, Skorokhodov, Siarohin, Menapace, Qian, Vasilkovsky, Lee, Wang, Zou, Tagliasacchi, et~al.]{bahmani2024vd3d}
Sherwin Bahmani, Ivan Skorokhodov, Aliaksandr Siarohin, Willi Menapace, Guocheng Qian, Michael Vasilkovsky, Hsin-Ying Lee, Chaoyang Wang, Jiaxu Zou, Andrea Tagliasacchi, et~al.
\newblock Vd3d: Taming large video diffusion transformers for 3d camera control.
\newblock \emph{arXiv preprint arXiv:2407.12781}, 2024{\natexlab{b}}.

\bibitem[Bai et~al.(2025)Bai, Xia, Fu, Wang, Mu, Cao, Liu, Hu, Bai, Wan, et~al.]{bai2025recammaster}
Jianhong Bai, Menghan Xia, Xiao Fu, Xintao Wang, Lianrui Mu, Jinwen Cao, Zuozhu Liu, Haoji Hu, Xiang Bai, Pengfei Wan, et~al.
\newblock Recammaster: Camera-controlled generative rendering from a single video.
\newblock \emph{arXiv preprint arXiv:2503.11647}, 2025.

\bibitem[Black et~al.(2023)Black, Janner, Du, Kostrikov, and Levine]{black2023training}
Kevin Black, Michael Janner, Yilun Du, Ilya Kostrikov, and Sergey Levine.
\newblock Training diffusion models with reinforcement learning.
\newblock \emph{arXiv preprint arXiv:2305.13301}, 2023.

\bibitem[Blattmann et~al.(2023)Blattmann, Dockhorn, Kulal, Mendelevitch, Kilian, Lorenz, Levi, English, Voleti, Letts, et~al.]{blattmann2023stable}
Andreas Blattmann, Tim Dockhorn, Sumith Kulal, Daniel Mendelevitch, Maciej Kilian, Dominik Lorenz, Yam Levi, Zion English, Vikram Voleti, Adam Letts, et~al.
\newblock Stable video diffusion: Scaling latent video diffusion models to large datasets.
\newblock \emph{arXiv preprint arXiv:2311.15127}, 2023.

\bibitem[Chan et~al.(2023)Chan, Nagano, Chan, Bergman, Park, Levy, Aittala, De~Mello, Karras, and Wetzstein]{chan2023generative}
Eric~R Chan, Koki Nagano, Matthew~A Chan, Alexander~W Bergman, Jeong~Joon Park, Axel Levy, Miika Aittala, Shalini De~Mello, Tero Karras, and Gordon Wetzstein.
\newblock Generative novel view synthesis with 3d-aware diffusion models.
\newblock In \emph{Proceedings of the IEEE/CVF International Conference on Computer Vision (ICCV)}, 2023.

\bibitem[Chen et~al.(2025)Chen, Zhou, Zhao, Wang, Zhang, Huang, Sun, Wen, and Li]{chen2025flexworld}
Luxi Chen, Zihan Zhou, Min Zhao, Yikai Wang, Ge Zhang, Wenhao Huang, Hao Sun, Ji-Rong Wen, and Chongxuan Li.
\newblock Flexworld: Progressively expanding 3d scenes for flexiable-view synthesis.
\newblock \emph{arXiv preprint arXiv:2503.13265}, 2025.

\bibitem[Duan et~al.(2025)Duan, Yu, Chen, Fei-Fei, and Wu]{duan2025worldscore}
Haoyi Duan, Hong-Xing Yu, Sirui Chen, Li Fei-Fei, and Jiajun Wu.
\newblock Worldscore: A unified evaluation benchmark for world generation.
\newblock \emph{arXiv preprint arXiv:2504.00983}, 2025.

\bibitem[Gao et~al.(2024)Gao, Holynski, Henzler, Brussee, Martin-Brualla, Srinivasan, Barron, and Poole]{gao2024cat3d}
Ruiqi Gao, Aleksander Holynski, Philipp Henzler, Arthur Brussee, Ricardo Martin-Brualla, Pratul Srinivasan, Jonathan~T Barron, and Ben Poole.
\newblock Cat3d: Create anything in 3d with multi-view diffusion models.
\newblock \emph{arXiv preprint arXiv:2405.10314}, 2024.

\bibitem[Ge et~al.(2023)Ge, Hu, Zhao, Liu, and Chen]{ge2023ref}
Wenhang Ge, Tao Hu, Haoyu Zhao, Shu Liu, and Ying-Cong Chen.
\newblock Ref-neus: Ambiguity-reduced neural implicit surface learning for multi-view reconstruction with reflection.
\newblock In \emph{Proceedings of the IEEE/CVF International Conference on Computer Vision (ICCV)}, 2023.

\bibitem[Ge et~al.(2024)Ge, Lin, Shen, Feng, Hu, Xu, and Chen]{ge2024prm}
Wenhang Ge, Jiantao Lin, Guibao Shen, Jiawei Feng, Tao Hu, Xinli Xu, and Ying-Cong Chen.
\newblock Prm: Photometric stereo based large reconstruction model.
\newblock 2024.

\bibitem[Grattafiori et~al.(2024)Grattafiori, Dubey, Jauhri, Pandey, Kadian, Al-Dahle, Letman, Mathur, Schelten, Vaughan, et~al.]{grattafiori2024llama}
Aaron Grattafiori, Abhimanyu Dubey, Abhinav Jauhri, Abhinav Pandey, Abhishek Kadian, Ahmad Al-Dahle, Aiesha Letman, Akhil Mathur, Alan Schelten, Alex Vaughan, et~al.
\newblock The llama 3 herd of models.
\newblock \emph{arXiv preprint arXiv:2407.21783}, 2024.

\bibitem[Gregory(2018)]{gregory2018game}
Jason Gregory.
\newblock \emph{Game engine architecture}.
\newblock AK Peters/CRC Press, 2018.

\bibitem[He et~al.(2024)He, Xu, Guo, Wetzstein, Dai, Li, and Yang]{he2024cameractrl}
Hao He, Yinghao Xu, Yuwei Guo, Gordon Wetzstein, Bo Dai, Hongsheng Li, and Ceyuan Yang.
\newblock Cameractrl: Enabling camera control for text-to-video generation.
\newblock \emph{arXiv preprint arXiv:2404.02101}, 2024.

\bibitem[He et~al.(2025)He, Yang, Lin, Xu, Wei, Gui, Zhao, Wetzstein, Jiang, and Li]{he2025cameractrl}
Hao He, Ceyuan Yang, Shanchuan Lin, Yinghao Xu, Meng Wei, Liangke Gui, Qi Zhao, Gordon Wetzstein, Lu Jiang, and Hongsheng Li.
\newblock Cameractrl ii: Dynamic scene exploration via camera-controlled video diffusion models.
\newblock \emph{arXiv preprint arXiv:2503.10592}, 2025.

\bibitem[Heusel et~al.(2017)Heusel, Ramsauer, Unterthiner, Nessler, and Hochreiter]{heusel2017gans}
Martin Heusel, Hubert Ramsauer, Thomas Unterthiner, Bernhard Nessler, and Sepp Hochreiter.
\newblock Gans trained by a two time-scale update rule converge to a local nash equilibrium.
\newblock \emph{Advances in neural information processing systems (NeurIPS)}, 2017.

\bibitem[Hong et~al.(2023)Hong, Zhang, Gu, Bi, Zhou, Liu, Liu, Sunkavalli, Bui, and Tan]{hong2023lrm}
Yicong Hong, Kai Zhang, Jiuxiang Gu, Sai Bi, Yang Zhou, Difan Liu, Feng Liu, Kalyan Sunkavalli, Trung Bui, and Hao Tan.
\newblock Lrm: Large reconstruction model for single image to 3d.
\newblock \emph{arXiv preprint arXiv:2311.04400}, 2023.

\bibitem[Hu et~al.(2025)Hu, Peng, Liu, and Ma]{hu2025ex}
Tao Hu, Haoyang Peng, Xiao Liu, and Yuewen Ma.
\newblock Ex-4d: Extreme viewpoint 4d video synthesis via depth watertight mesh.
\newblock \emph{arXiv preprint arXiv:2506.05554}, 2025.

\bibitem[Jiang et~al.(2025)Jiang, Lin, Chen, Ge, Yang, Jiang, Lyu, Zheng, and Chen]{jiang2025dimer}
Lutao Jiang, Jiantao Lin, Kanghao Chen, Wenhang Ge, Xin Yang, Yifan Jiang, Yuanhuiyi Lyu, Xu Zheng, and Yingcong Chen.
\newblock Dimer: Disentangled mesh reconstruction model.
\newblock \emph{arXiv preprint arXiv:2504.17670}, 2025.

\bibitem[Kerbl et~al.(2023)Kerbl, Kopanas, Leimk{\"u}hler, and Drettakis]{kerbl3Dgaussians}
Bernhard Kerbl, Georgios Kopanas, Thomas Leimk{\"u}hler, and George Drettakis.
\newblock 3d gaussian splatting for real-time radiance field rendering.
\newblock \emph{ACM Transactions on Graphics (ToG)}, 2023.

\bibitem[Kingma and Ba(2014)]{kingma2014adam}
Diederik~P Kingma and Jimmy Ba.
\newblock Adam: A method for stochastic optimization.
\newblock \emph{arXiv preprint arXiv:1412.6980}, 2014.

\bibitem[Lee et~al.(2023)Lee, Phatale, Mansoor, Lu, Mesnard, Ferret, Bishop, Hall, Carbune, and Rastogi]{lee2023rlaif}
Harrison Lee, Samrat Phatale, Hassan Mansoor, Kellie~Ren Lu, Thomas Mesnard, Johan Ferret, Colton Bishop, Ethan Hall, Victor Carbune, and Abhinav Rastogi.
\newblock Rlaif: Scaling reinforcement learning from human feedback with ai feedback.
\newblock 2023.

\bibitem[Li(2025)]{li2025worldlabs}
Fei-Fei Li.
\newblock Worldlabs, 2025.

\bibitem[Li et~al.(2024{\natexlab{a}})Li, Feng, Fu, Wang, Basu, Chen, and Wang]{li2024t2v}
Jiachen Li, Weixi Feng, Tsu-Jui Fu, Xinyi Wang, Sugato Basu, Wenhu Chen, and William~Yang Wang.
\newblock T2v-turbo: Breaking the quality bottleneck of video consistency model with mixed reward feedback.
\newblock \emph{arXiv preprint arXiv:2405.18750}, 2024{\natexlab{a}}.

\bibitem[Li et~al.(2024{\natexlab{b}})Li, Yang, Kuang, Wu, Wang, Xiao, and Chen]{li2024controlnet++}
Ming Li, Taojiannan Yang, Huafeng Kuang, Jie Wu, Zhaoning Wang, Xuefeng Xiao, and Chen Chen.
\newblock Controlnet++: Improving conditional controls with efficient consistency feedback: Project page: liming-ai. github. io/controlnet\_plus\_plus.
\newblock In \emph{European Conference on Computer Vision (ECCV)}, 2024{\natexlab{b}}.

\bibitem[Li et~al.(2025)Li, Zheng, Jiang, Zhan, Wu, Lu, Lin, Deng, Xiong, Chen, et~al.]{li2025realcam}
Teng Li, Guangcong Zheng, Rui Jiang, Shuigen Zhan, Tao Wu, Yehao Lu, Yining Lin, Chuanyun Deng, Yepan Xiong, Min Chen, et~al.
\newblock Realcam-i2v: Real-world image-to-video generation with interactive complex camera control.
\newblock In \emph{Proceedings of the IEEE/CVF International Conference on Computer Vision (ICCV)}, 2025.

\bibitem[Liang et~al.(2024)Liang, Cao, Goel, Qian, Korolev, Terzopoulos, Plataniotis, Tulyakov, and Ren]{liang2024wonderland}
Hanwen Liang, Junli Cao, Vidit Goel, Guocheng Qian, Sergei Korolev, Demetri Terzopoulos, Konstantinos Plataniotis, Sergey Tulyakov, and Jian Ren.
\newblock Wonderland: Navigating 3d scenes from a single image.
\newblock \emph{arXiv preprint arXiv:2412.12091}, 2024.

\bibitem[Liu et~al.(2025)Liu, Liu, Liang, Yuan, Liu, Zheng, Wu, Wang, Qin, Xia, et~al.]{liu2025improving}
Jie Liu, Gongye Liu, Jiajun Liang, Ziyang Yuan, Xiaokun Liu, Mingwu Zheng, Xiele Wu, Qiulin Wang, Wenyu Qin, Menghan Xia, et~al.
\newblock Improving video generation with human feedback.
\newblock \emph{arXiv preprint arXiv:2501.13918}, 2025.

\bibitem[Liu et~al.(2024)Liu, Zhang, Li, Yan, Gao, Chen, Yuan, Huang, Sun, Gao, et~al.]{liu2024sora}
Yixin Liu, Kai Zhang, Yuan Li, Zhiling Yan, Chujie Gao, Ruoxi Chen, Zhengqing Yuan, Yue Huang, Hanchi Sun, Jianfeng Gao, et~al.
\newblock Sora: A review on background, technology, limitations, and opportunities of large vision models.
\newblock \emph{arXiv preprint arXiv:2402.17177}, 2024.

\bibitem[Mateo et~al.(2016)Mateo, Gil, and Torres]{mateo2016visual}
CM Mateo, P Gil, and F Torres.
\newblock Visual perception for the 3d recognition of geometric pieces in robotic manipulation.
\newblock \emph{The International Journal of Advanced Manufacturing Technology}, 2016.

\bibitem[Peebles and Xie(2022)]{Peebles2022DiT}
William Peebles and Saining Xie.
\newblock Scalable diffusion models with transformers.
\newblock \emph{arXiv preprint arXiv:2212.09748}, 2022.

\bibitem[Prabhudesai et~al.(2023)Prabhudesai, Goyal, Pathak, and Fragkiadaki]{prabhudesai2023aligning}
Mihir Prabhudesai, Anirudh Goyal, Deepak Pathak, and Katerina Fragkiadaki.
\newblock Aligning text-to-image diffusion models with reward backpropagation.
\newblock 2023.

\bibitem[Prabhudesai et~al.(2024)Prabhudesai, Mendonca, Qin, Fragkiadaki, and Pathak]{prabhudesai2024video}
Mihir Prabhudesai, Russell Mendonca, Zheyang Qin, Katerina Fragkiadaki, and Deepak Pathak.
\newblock Video diffusion alignment via reward gradients.
\newblock \emph{arXiv preprint arXiv:2407.08737}, 2024.

\bibitem[Ren et~al.(2025)Ren, Shen, Huang, Ling, Lu, Nimier-David, M{\"u}ller, Keller, Fidler, and Gao]{ren2025gen3c}
Xuanchi Ren, Tianchang Shen, Jiahui Huang, Huan Ling, Yifan Lu, Merlin Nimier-David, Thomas M{\"u}ller, Alexander Keller, Sanja Fidler, and Jun Gao.
\newblock Gen3c: 3d-informed world-consistent video generation with precise camera control.
\newblock \emph{arXiv preprint arXiv:2503.03751}, 2025.

\bibitem[Sargent et~al.(2024)Sargent, Li, Shah, Herrmann, Yu, Zhang, Chan, Lagun, Fei-Fei, Sun, et~al.]{sargent2024zeronvs}
Kyle Sargent, Zizhang Li, Tanmay Shah, Charles Herrmann, Hong-Xing Yu, Yunzhi Zhang, Eric~Ryan Chan, Dmitry Lagun, Li Fei-Fei, Deqing Sun, et~al.
\newblock Zeronvs: Zero-shot 360-degree view synthesis from a single image.
\newblock In \emph{Proceedings of the IEEE/CVF Conference on Computer Vision and Pattern Recognition (CVPR)}, 2024.

\bibitem[Sch\"{o}nberger and Frahm(2016)]{schoenberger2016sfm}
Johannes~Lutz Sch\"{o}nberger and Jan-Michael Frahm.
\newblock Structure-from-motion revisited.
\newblock In \emph{Conference on Computer Vision and Pattern Recognition (CVPR)}, 2016.

\bibitem[Schuemie et~al.(2001)Schuemie, Van Der~Straaten, Krijn, and Van Der~Mast]{schuemie2001research}
Martijn~J Schuemie, Peter Van Der~Straaten, Merel Krijn, and Charles~APG Van Der~Mast.
\newblock Research on presence in virtual reality: A survey.
\newblock \emph{Cyberpsychology \& behavior}, 2001.

\bibitem[Sitzmann et~al.(2021)Sitzmann, Rezchikov, Freeman, Tenenbaum, and Durand]{sitzmann2021light}
Vincent Sitzmann, Semon Rezchikov, Bill Freeman, Josh Tenenbaum, and Fredo Durand.
\newblock Light field networks: Neural scene representations with single-evaluation rendering.
\newblock \emph{Advances in Neural Information Processing Systems (NeurIPS)}, 2021.

\bibitem[Skalse et~al.(2022)Skalse, Howe, Krasheninnikov, and Krueger]{skalse2022defining}
Joar Skalse, Nikolaus Howe, Dmitrii Krasheninnikov, and David Krueger.
\newblock Defining and characterizing reward gaming.
\newblock \emph{Advances in Neural Information Processing Systems (NeurIPS)}, 2022.

\bibitem[Sun et~al.(2024)Sun, Chen, Liu, Chen, Duan, Zhang, and Wang]{sun2024dimensionx}
Wenqiang Sun, Shuo Chen, Fangfu Liu, Zilong Chen, Yueqi Duan, Jun Zhang, and Yikai Wang.
\newblock Dimensionx: Create any 3d and 4d scenes from a single image with controllable video diffusion.
\newblock \emph{arXiv preprint arXiv:2411.04928}, 2024.

\bibitem[Teed and Deng(2021)]{teed2021droid}
Zachary Teed and Jia Deng.
\newblock Droid-slam: Deep visual slam for monocular, stereo, and rgb-d cameras.
\newblock \emph{Advances in neural information processing systems (NeurIPS)}, 2021.

\bibitem[Unterthiner et~al.(2019)Unterthiner, Van~Steenkiste, Kurach, Marinier, Michalski, and Gelly]{unterthiner2019fvd}
Thomas Unterthiner, Sjoerd Van~Steenkiste, Karol Kurach, Rapha{\"e}l Marinier, Marcin Michalski, and Sylvain Gelly.
\newblock Fvd: A new metric for video generation.
\newblock 2019.

\bibitem[Voleti et~al.(2024)Voleti, Yao, Boss, Letts, Pankratz, Tochilkin, Laforte, Rombach, and Jampani]{voleti2024sv3d}
Vikram Voleti, Chun-Han Yao, Mark Boss, Adam Letts, David Pankratz, Dmitry Tochilkin, Christian Laforte, Robin Rombach, and Varun Jampani.
\newblock Sv3d: Novel multi-view synthesis and 3d generation from a single image using latent video diffusion.
\newblock In \emph{European Conference on Computer Vision (ECCV)}, 2024.

\bibitem[Wallace et~al.(2024)Wallace, Dang, Rafailov, Zhou, Lou, Purushwalkam, Ermon, Xiong, Joty, and Naik]{wallace2024diffusion}
Bram Wallace, Meihua Dang, Rafael Rafailov, Linqi Zhou, Aaron Lou, Senthil Purushwalkam, Stefano Ermon, Caiming Xiong, Shafiq Joty, and Nikhil Naik.
\newblock Diffusion model alignment using direct preference optimization.
\newblock In \emph{Proceedings of the IEEE/CVF Conference on Computer Vision and Pattern Recognition (CVPR)}, 2024.

\bibitem[Wang et~al.(2025)Wang, Chen, Karaev, Vedaldi, Rupprecht, and Novotny]{wang2025vggt}
Jianyuan Wang, Minghao Chen, Nikita Karaev, Andrea Vedaldi, Christian Rupprecht, and David Novotny.
\newblock Vggt: Visual geometry grounded transformer.
\newblock In \emph{Proceedings of the IEEE/CVF Conference on Computer Vision and Pattern Recognition (CVPR)}, 2025.

\bibitem[Wang et~al.(2024{\natexlab{a}})Wang, Leroy, Cabon, Chidlovskii, and Revaud]{wang2024dust3r}
Shuzhe Wang, Vincent Leroy, Yohann Cabon, Boris Chidlovskii, and Jerome Revaud.
\newblock Dust3r: Geometric 3d vision made easy.
\newblock In \emph{Proceedings of the IEEE/CVF Conference on Computer Vision and Pattern Recognition (CVPR)}, 2024{\natexlab{a}}.

\bibitem[Wang et~al.(2004)Wang, Bovik, Sheikh, and Simoncelli]{wang2004image}
Zhou Wang, Alan~C Bovik, Hamid~R Sheikh, and Eero~P Simoncelli.
\newblock Image quality assessment: from error visibility to structural similarity.
\newblock \emph{IEEE transactions on image processing (TIP)}, 2004.

\bibitem[Wang et~al.(2024{\natexlab{b}})Wang, Yuan, Wang, Li, Chen, Xia, Luo, and Shan]{wang2024motionctrl}
Zhouxia Wang, Ziyang Yuan, Xintao Wang, Yaowei Li, Tianshui Chen, Menghan Xia, Ping Luo, and Ying Shan.
\newblock Motionctrl: A unified and flexible motion controller for video generation.
\newblock In \emph{ACM SIGGRAPH 2024 Conference Papers (SIGGRAPH)}, 2024{\natexlab{b}}.

\bibitem[Wu et~al.(2025)Wu, Gao, Poole, Trevithick, Zheng, Barron, and Holynski]{wu2025cat4d}
Rundi Wu, Ruiqi Gao, Ben Poole, Alex Trevithick, Changxi Zheng, Jonathan~T Barron, and Aleksander Holynski.
\newblock Cat4d: Create anything in 4d with multi-view video diffusion models.
\newblock In \emph{Proceedings of the IEEE/CVF Conference on Computer Vision and Pattern Recognition (CVPR)}, 2025.

\bibitem[Xu et~al.(2024{\natexlab{a}})Xu, Nie, Liu, Liu, Kautz, Wang, and Vahdat]{xu2024camco}
Dejia Xu, Weili Nie, Chao Liu, Sifei Liu, Jan Kautz, Zhangyang Wang, and Arash Vahdat.
\newblock Camco: Camera-controllable 3d-consistent image-to-video generation.
\newblock \emph{arXiv preprint arXiv:2406.02509}, 2024{\natexlab{a}}.

\bibitem[Xu et~al.(2023)Xu, Liu, Wu, Tong, Li, Ding, Tang, and Dong]{xu2023imagereward}
Jiazheng Xu, Xiao Liu, Yuchen Wu, Yuxuan Tong, Qinkai Li, Ming Ding, Jie Tang, and Yuxiao Dong.
\newblock Imagereward: Learning and evaluating human preferences for text-to-image generation.
\newblock \emph{Advances in Neural Information Processing Systems (NeurIPS)}, 2023.

\bibitem[Xu et~al.(2024{\natexlab{b}})Xu, Ge, Lin, Feng, Xu, Zhao, Zhang, and Chen]{xu2024flexgen}
Xinli Xu, Wenhang Ge, Jiantao Lin, Jiawei Feng, Lie Xu, HanFeng Zhao, Shunsi Zhang, and Ying-Cong Chen.
\newblock Flexgen: Flexible multi-view generation from text and image inputs.
\newblock \emph{arXiv preprint}, 2024{\natexlab{b}}.

\bibitem[Yang et~al.(2024{\natexlab{a}})Yang, Yang, Zhang, Hui, Zheng, Yu, Li, Liu, Huang, Wei, et~al.]{yang2024qwen2}
An Yang, Baosong Yang, Beichen Zhang, Binyuan Hui, Bo Zheng, Bowen Yu, Chengyuan Li, Dayiheng Liu, Fei Huang, Haoran Wei, et~al.
\newblock Qwen2. 5 technical report.
\newblock \emph{arXiv preprint arXiv:2412.15115}, 2024{\natexlab{a}}.

\bibitem[Yang et~al.(2024{\natexlab{b}})Yang, Tao, Lyu, Ge, Chen, Shen, Zhu, and Li]{yang2024using}
Kai Yang, Jian Tao, Jiafei Lyu, Chunjiang Ge, Jiaxin Chen, Weihan Shen, Xiaolong Zhu, and Xiu Li.
\newblock Using human feedback to fine-tune diffusion models without any reward model.
\newblock In \emph{Proceedings of the IEEE/CVF Conference on Computer Vision and Pattern Recognition (CVPR)}, 2024{\natexlab{b}}.

\bibitem[Yang et~al.(2025)Yang, Xu, Luan, Zhan, Qiu, Shi, Li, Yang, Zhang, Yu, et~al.]{yang2025omnicam}
Xiaoda Yang, Jiayang Xu, Kaixuan Luan, Xinyu Zhan, Hongshun Qiu, Shijun Shi, Hao Li, Shuai Yang, Li Zhang, Checheng Yu, et~al.
\newblock Omnicam: Unified multimodal video generation via camera control.
\newblock \emph{arXiv preprint arXiv:2504.02312}, 2025.

\bibitem[Yang et~al.(2024{\natexlab{c}})Yang, Teng, Zheng, Ding, Huang, Xu, Yang, Hong, Zhang, Feng, et~al.]{yang2024cogvideox}
Zhuoyi Yang, Jiayan Teng, Wendi Zheng, Ming Ding, Shiyu Huang, Jiazheng Xu, Yuanming Yang, Wenyi Hong, Xiaohan Zhang, Guanyu Feng, et~al.
\newblock Cogvideox: Text-to-video diffusion models with an expert transformer.
\newblock \emph{arXiv preprint arXiv:2408.06072}, 2024{\natexlab{c}}.

\bibitem[Yu et~al.(2024)Yu, Xing, Yuan, Hu, Li, Huang, Gao, Wong, Shan, and Tian]{yu2024viewcrafter}
Wangbo Yu, Jinbo Xing, Li Yuan, Wenbo Hu, Xiaoyu Li, Zhipeng Huang, Xiangjun Gao, Tien-Tsin Wong, Ying Shan, and Yonghong Tian.
\newblock Viewcrafter: Taming video diffusion models for high-fidelity novel view synthesis.
\newblock \emph{arXiv preprint arXiv:2409.02048}, 2024.

\bibitem[Yuan et~al.(2024)Yuan, Zhang, Wang, Wei, Feng, Pan, Zhang, Liu, Albanie, and Ni]{yuan2024instructvideo}
Hangjie Yuan, Shiwei Zhang, Xiang Wang, Yujie Wei, Tao Feng, Yining Pan, Yingya Zhang, Ziwei Liu, Samuel Albanie, and Dong Ni.
\newblock Instructvideo: Instructing video diffusion models with human feedback.
\newblock In \emph{Proceedings of the IEEE/CVF Conference on Computer Vision and Pattern Recognition (CVPR)}, 2024.

\bibitem[Zhang et~al.(2024{\natexlab{a}})Zhang, Wu, Ren, Xia, Kuang, Xie, Li, Xiao, Huang, Wen, et~al.]{zhang2024unifl}
Jiacheng Zhang, Jie Wu, Yuxi Ren, Xin Xia, Huafeng Kuang, Pan Xie, Jiashi Li, Xuefeng Xiao, Weilin Huang, Shilei Wen, et~al.
\newblock Unifl: Improve latent diffusion model via unified feedback learning.
\newblock \emph{Advances in Neural Information Processing (NeurIPS)}, 2024{\natexlab{a}}.

\bibitem[Zhang et~al.(2024{\natexlab{b}})Zhang, Bi, Tan, Xiangli, Zhao, Sunkavalli, and Xu]{GS-LRM}
Kai Zhang, Sai Bi, Hao Tan, Yuanbo Xiangli, Nanxuan Zhao, Kalyan Sunkavalli, and Zexiang Xu.
\newblock Gs-lrm: Large reconstruction model for 3d gaussian splatting.
\newblock In \emph{European Conference on Computer Vision (ECCV)}, 2024{\natexlab{b}}.

\bibitem[Zhang et~al.(2023)Zhang, Rao, and Agrawala]{zhang2023adding}
Lvmin Zhang, Anyi Rao, and Maneesh Agrawala.
\newblock Adding conditional control to text-to-image diffusion models.
\newblock In \emph{Proceedings of the IEEE/CVF international conference on computer vision (ICCV)}, 2023.

\bibitem[Zhang et~al.(2024{\natexlab{c}})Zhang, Wang, Zhang, Qiu, Pang, Jiang, Yang, Xu, and Yu]{zhang2024clay}
Longwen Zhang, Ziyu Wang, Qixuan Zhang, Qiwei Qiu, Anqi Pang, Haoran Jiang, Wei Yang, Lan Xu, and Jingyi Yu.
\newblock Clay: A controllable large-scale generative model for creating high-quality 3d assets.
\newblock \emph{arXiv preprint arXiv:2406.13897}, 2024{\natexlab{c}}.

\bibitem[Zhang et~al.(2018{\natexlab{a}})Zhang, Isola, Efros, Shechtman, and Wang]{lipis}
Richard Zhang, Phillip Isola, Alexei~A Efros, Eli Shechtman, and Oliver Wang.
\newblock The unreasonable effectiveness of deep features as a perceptual metric.
\newblock In \emph{Proceedings of the IEEE conference on computer vision and pattern recognition (CVPR)}, 2018{\natexlab{a}}.

\bibitem[Zhang et~al.(2018{\natexlab{b}})Zhang, Isola, Efros, Shechtman, and Wang]{zhang2018unreasonable}
Richard Zhang, Phillip Isola, Alexei~A Efros, Eli Shechtman, and Oliver Wang.
\newblock The unreasonable effectiveness of deep features as a perceptual metric.
\newblock In \emph{Proceedings of the IEEE conference on computer vision and pattern recognition (CVPR)}, 2018{\natexlab{b}}.

\bibitem[Zheng et~al.(2024)Zheng, Li, Jiang, Lu, Wu, and Li]{zheng2024cami2v}
Guangcong Zheng, Teng Li, Rui Jiang, Yehao Lu, Tao Wu, and Xi Li.
\newblock Cami2v: Camera-controlled image-to-video diffusion model.
\newblock \emph{arXiv preprint arXiv:2410.15957}, 2024.

\bibitem[Zhou et~al.(2018)Zhou, Tucker, Flynn, Fyffe, and Snavely]{zhou2018stereo}
Tinghui Zhou, Richard Tucker, John Flynn, Graham Fyffe, and Noah Snavely.
\newblock Stereo magnification: Learning view synthesis using multiplane images.
\newblock \emph{arXiv preprint arXiv:1805.09817}, 2018.

\end{thebibliography}
}

\clearpage
\setcounter{page}{1}
\maketitlesupplementary

\section{Further Analysis and discussion}

\noindent\textbf{Further Discussion on the Improvements of Using ReFL.}
We discuss more improvements of after ReFL.
We visualized the qualitative comparison in the main text.
From the first case, we can observe that ``w/ ReFL'' maintains better photometric consistency during camera motion. In the results ``w/o ReFL,'' there is an obvious photometric shift. 
Our camera-aware 3D decoder leverages 3DGS to represent the scene, which is typically photometrically consistent across novel views. 
This property is also distilled into the video diffusion model by ReFL, which is favorable for this task.
Moreover, we found that ``w/ ReFL'' can effectively suppress dynamic generation, maintaining better 3D consistency in generated videos. Since 3DGS is essentially a static 3D representation, this property is also distilled into the video model to produce content that is both static and 3D consistent. The corresponding video can be found in the Supplementary Materials.

\begin{figure*}
    \centering
    \includegraphics[width=0.9\linewidth]{ 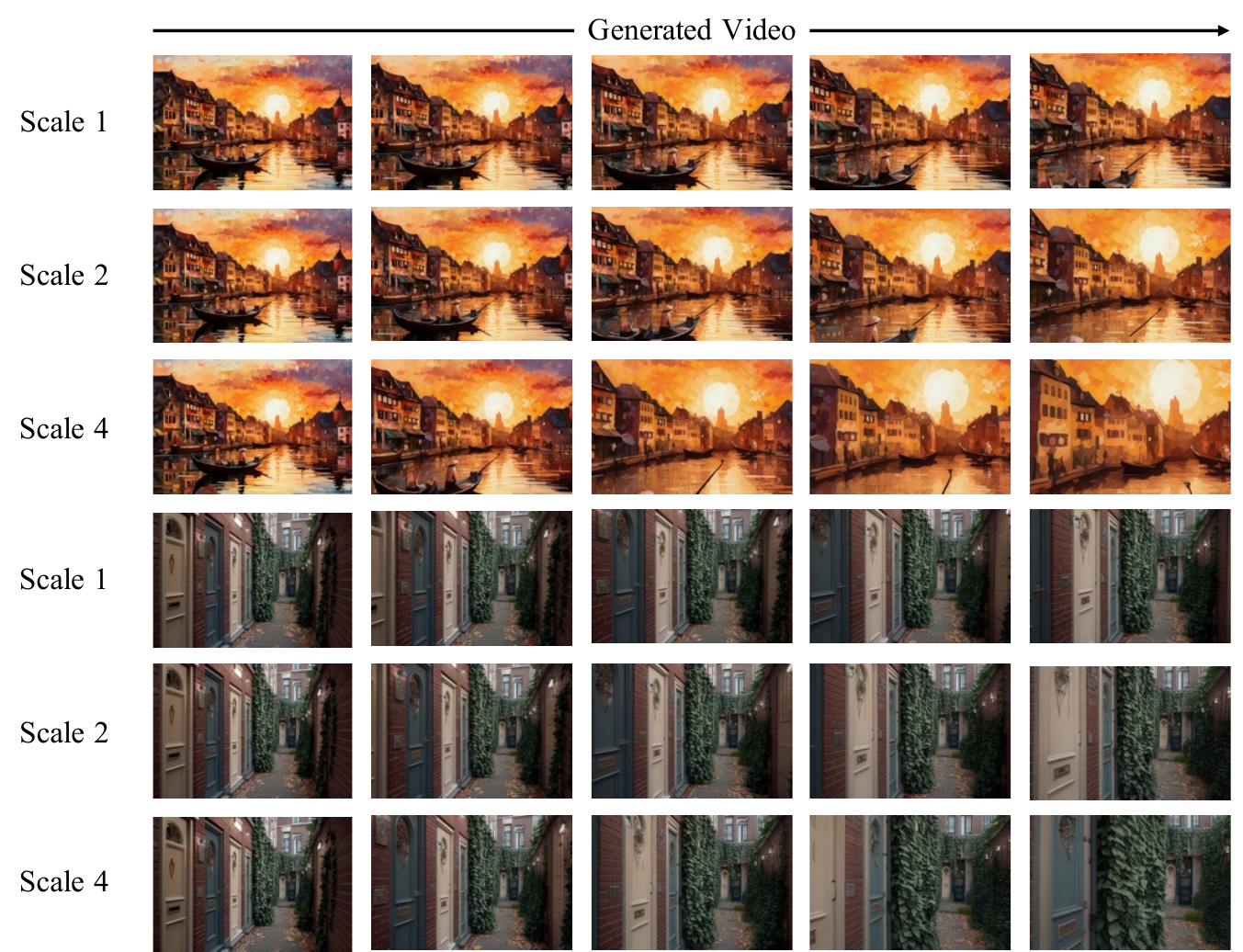}
    \caption{Our model is capable of perceiving scale variations and generating videos that accurately reflect the intended degree of movement. A larger scale results in more pronounced movements.}
    \label{fig:scale}
\end{figure*}

\vspace{0.1cm}
\noindent\textbf{The scale of camera conditions}.
Although the camera poses in the RE10K dataset are normalized to a unified scale as described in \citep{zhou2018stereo}, we observed that there are still variations in scale within this unified framework. Specifically, some movements are more pronounced while others are subtler. During inference, we found that by manipulating the scale of the camera conditions, our model can effectively perceive these scale variations and generate videos that accurately reflect the intended degree of movement. We visualized some examples in Fig. \ref{fig:scale}, 
where the same image was used as a condition, but the scale of the camera pose was varied for each generation. 

\begin{figure*}
    \centering
    \includegraphics[width=0.85\linewidth]{ 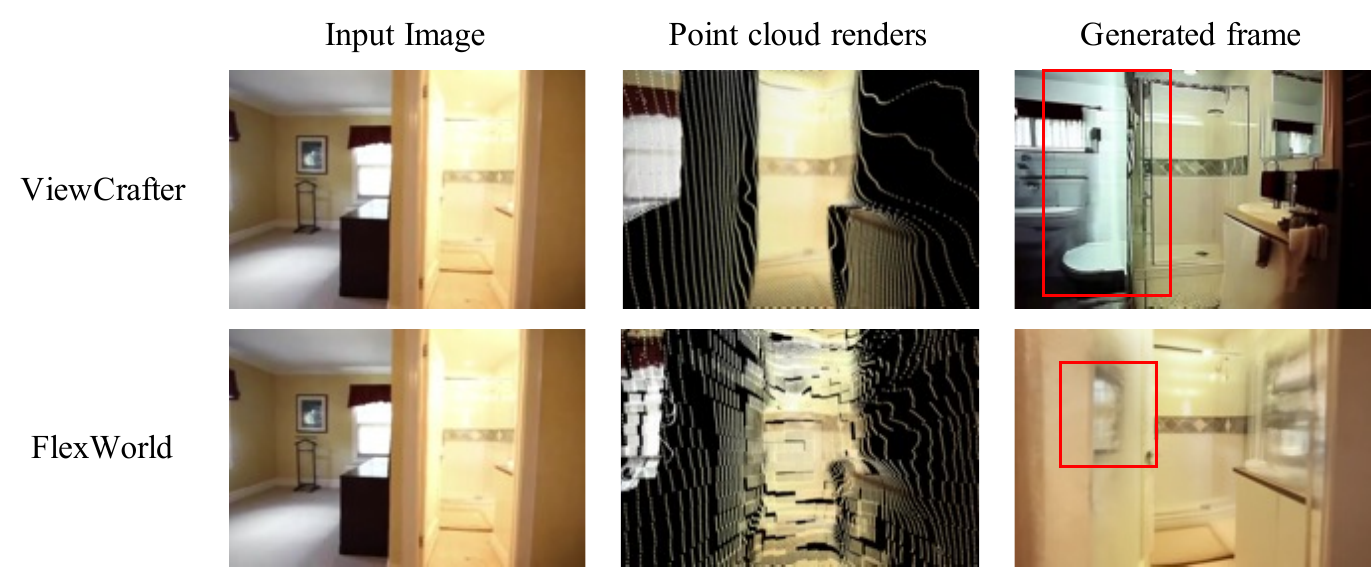}
    \caption{Using point cloud renders as camera condition incurs a rendering leakage problem, affecting the quality of novel view synthesis.}
    \label{fig:pointrender}
\end{figure*}

\vspace{0.1cm}
\noindent\textbf{The choice of Plücker embeddings as conditions}.
Recent camera-conditioned video generation methods can be roughly divided into two categories: those that use point cloud renders as conditions and those that use Plücker embeddings as conditions. We chose Plücker embeddings due to their flexibility and generalization capabilities. However, our method is general and can also be employed in frameworks where point cloud renders are used as conditions.
Using point cloud renders as conditions typically relies on external models \citep{wang2024dust3r, wang2025vggt} for simultaneous point cloud and camera pose estimation to achieve alignment. If a dataset contains ground-truth metric camera poses, the estimated point cloud should be further processed to align with the ground-truth poses, while Plücker embeddings can be easily obtained without any preprocessing. 
Moreover, point cloud renders incur a rendering leakage problem: as the camera view changes, background points may be incorrectly rendered into the foreground due to improper handling of occlusion relationships, affecting the realism and consistency. We show an example in Fig. \ref{fig:pointrender}.

\begin{figure*}
    \centering
    \includegraphics[width=0.95\linewidth]{ 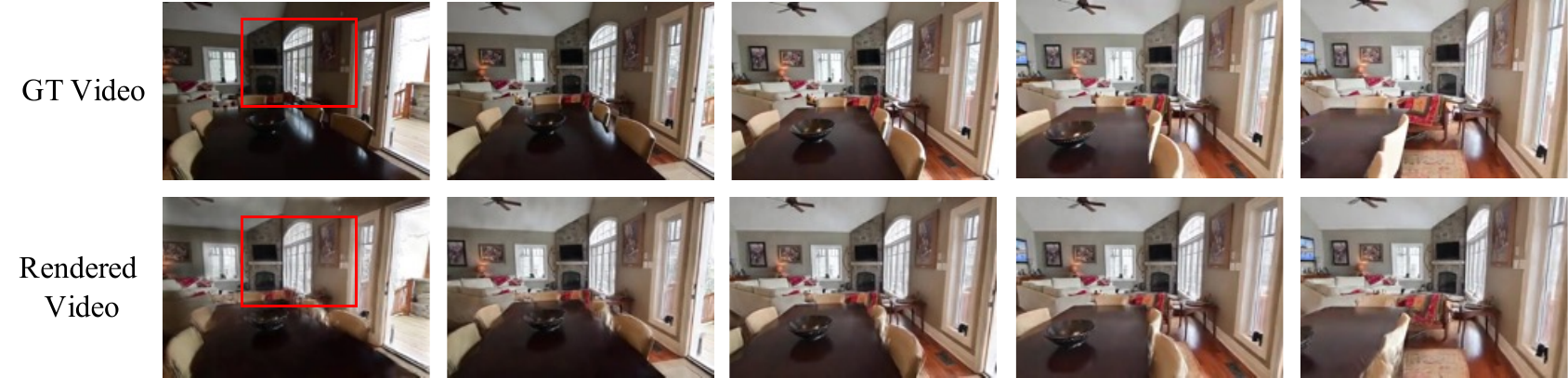}
    \caption{An example of ground-truth videos with varying exposure levels. The rendered video from 3DGS tends to exhibit an average exposure, which differs from the ground-truth video.}
    \label{fig:illumination}
\end{figure*}

\vspace{0.1cm}
\noindent\textbf{The reconstruction performance of camera-aware 3D decoder.} 
Our camera-aware 3D decoder is exclusively trained on the RE10K dataset, which comprises estate videos exhibiting varying exposure changes as the camera perspective shifts. The model generates per-frame 3DGS and uses them as a global 3D representation for rendering. However, exposure changes result in variations in the predicted spherical harmonics, which can degrade rendering quality to some extent. We show some examples in Fig. \ref{fig:illumination}.  Collecting more consistent videos with precise camera poses can further enhance the reconstruction performance of the camera-aware 3D decoder.

\section{The efficiency of camera-aware 3D decoder}
We compared the efficiency of our proposed camera-aware 3D decoder and video VAE decoder in terms of GPU memory cost and time cost, as shown in Table \ref{table:cost}. When using the video VAE decoder, we can only decode 2 temporal latents in each iteration with 80GB of GPU memory during ReFL training, while camera-aware 3D decoder can decoder much more frames. Moreover, the visibility mask is not available with video VAE decoder.

\begin{table}[h]
\centering
\footnotesize
\caption{Comparison of GPU Memory and Time Cost}
\label{table:cost}
\begin{tabular}{lcc}
\toprule
\textbf{Decoder Type} & \textbf{GPU Memory Cost (GB)} & \textbf{Time Cost (s)} \\ 
\midrule
Our Decoder & 8.44 & 0.559\\ 
Video VAE Decoder & 43.17 & 5.602 \\ 
\bottomrule
\end{tabular}
\end{table}

\section{Optimization and Additional Model Details}
\noindent \textbf{Optimization Details.} 
We used the Adam optimizer \citep{kingma2014adam}. In the first stage, the learning rate was set to \(1 \times 10^{-4}\). In the second stage, the learning rate was set to \(3 \times 10^{-4}\), and in the third fine-tuning stage, the learning rate was set to \(1 \times 10^{-5}\).
In the first stage, we trained the basic camera-controlled video model with a batch size of 16 for 10K steps. In the second stage, we  trained our camera-aware 3D decoder with a batch size of 32 for 100K steps.
In the third stage, we performed reward-based feedback learning with a batch size of 16 for 5K steps. 

\begin{figure*}
    \centering
    \begin{minipage}{0.45\linewidth}
        \centering
        \includegraphics[width=\linewidth]{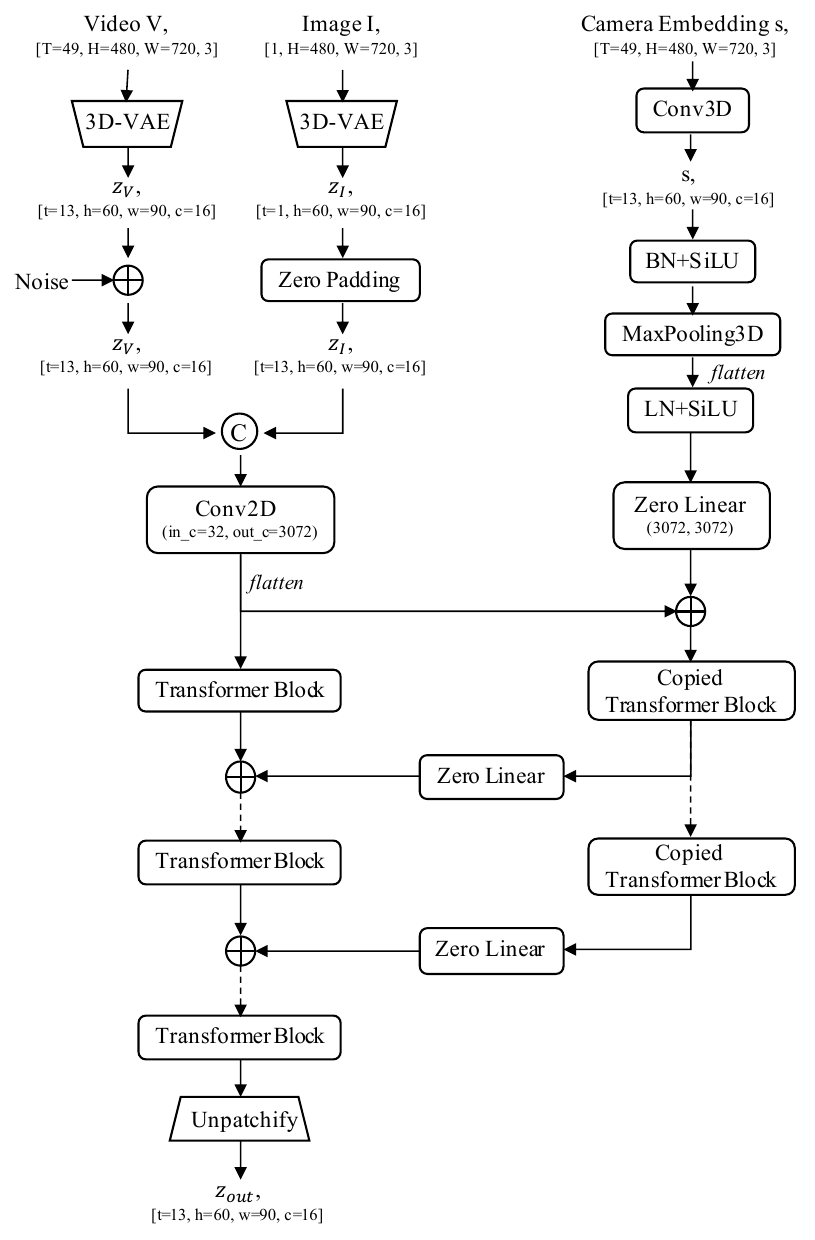}
        \caption{The detailed network architecture for camera-controlled video diffusion model.}
        \label{fig:video}
    \end{minipage}
    \hspace{0.05\linewidth} 
    \begin{minipage}{0.45\linewidth}
        \centering
        \includegraphics[width=0.85\linewidth]{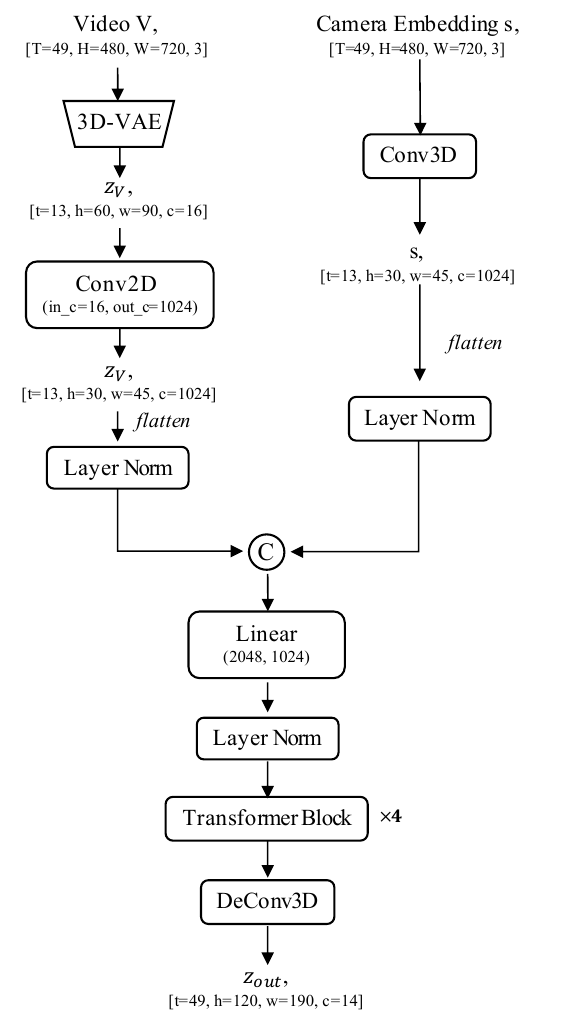} 
        \caption{The detailed network architecture for camera-aware 3D decoder.}
        \label{fig:decoder}
    \end{minipage}
\end{figure*}

\vspace{0.1cm}
\noindent \textbf{Network architecture.} 
 The details of the network for the first stage are shown in Fig. \ref{fig:video}. Pixel-aligned Plücker embeddings are compressed via a Conv3D layer, ensuring the camera latent shares the same dimension with the video latent. Then, batch normalization, an activation layer, and a max pooling layer are used to convert the camera latent into sequential tokens as ControlNet input. For efficiency considerations, we only copied the first 8 transformer blocks.

For the camera-aware 3D decoder, we elaborate on the network architecture in Fig. \ref{fig:decoder}. We convert the video latent using Conv2D into visual tokens. To ensure the same dimension for the camera embedding, we leverage Conv3D for spatial-temporal compression. Then, visual tokens and camera tokens are concatenated along the channel dimension. Four Transformer blocks and a DeConv3D layer are used to process the concatenated tokens into pixel-aligned 3DGS. Note that we do not recover the original spatial resolution for 3DGS, which we found is sufficient to represent a scene. During training, we employed 49 supervision views, where 14 frames are randomly sampled from the source video clip as seen views, and the remaining 35 are selected from disjoint frames as unseen views to ensure 3D consistency.

\section{Qualitative comparison on WorldScore static benchmark}
\begin{figure*}
    \centering
    \includegraphics[width=\linewidth]{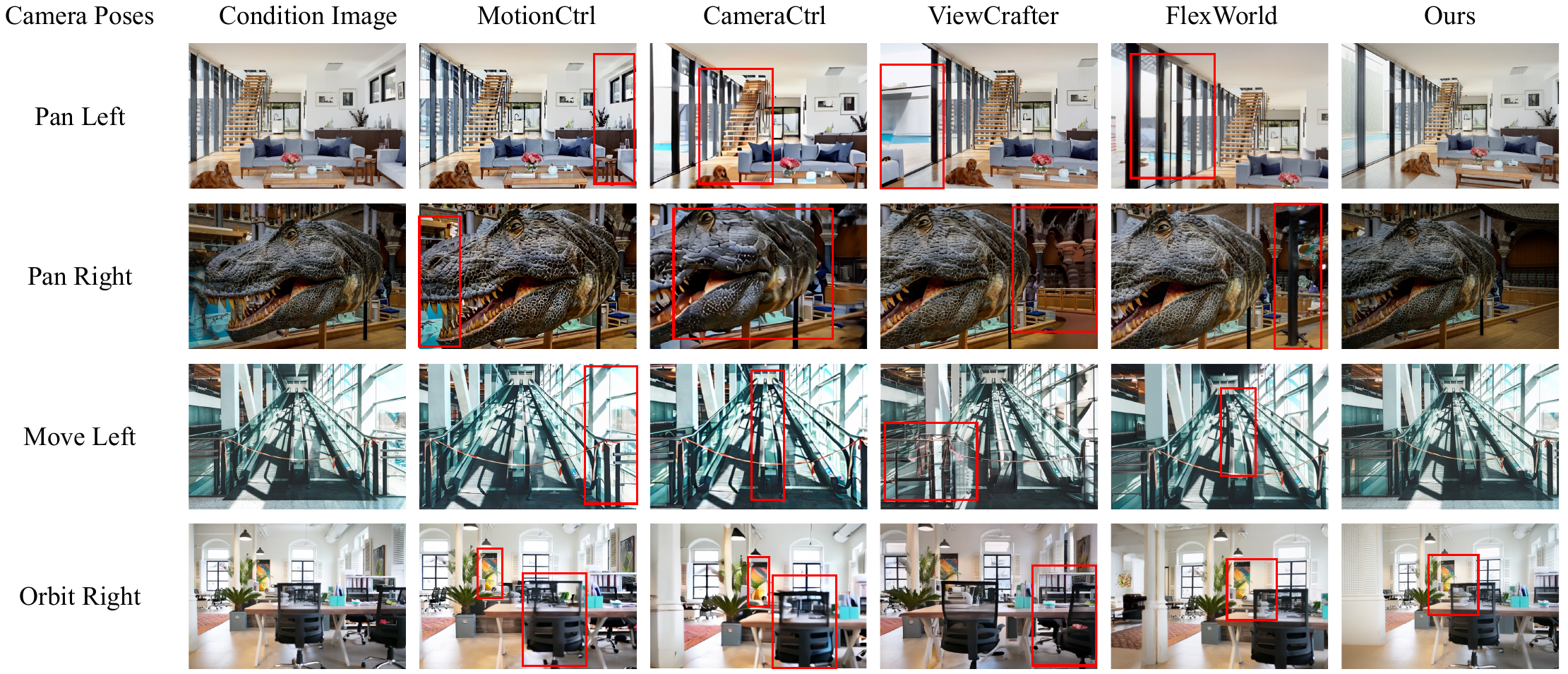}
    \caption{Qualitative comparison on WorldScore static benchmark.}
    \label{figws}
\end{figure*}

We further visualize the qualitative comparison on the WorldScore static benchmark in Fig. \ref{figws}. Our method generates more 3D consistent videos that match the given camera conditions.

\section{Plücker embeddings derivation}
Given a camera trajectory with extrinsic parameters $\mathbf{E} = [\mathbf{R}; \mathbf{t}] \in \mathbb{R}^{3 \times 4}$ and intrinsic matrix $\mathbf{K} \in \mathbb{R}^{3 \times 3}$, we derive the Plücker representation $\mathbf{s} = (\mathbf{o} \times \mathbf{d}', \mathbf{d}')$ for each pixel $(u, v)$. The camera's world-space origin $\mathbf{o}$ is defined by the translation vector $\mathbf{t}$. The direction vector $\mathbf{d}$ from the camera center to the pixel is computed as:
\[
\mathbf{d} = \mathbf{R} \mathbf{K}^{-1} [u, v, 1]^T + \mathbf{t},
\]
where $\mathbf{K}^{-1} [u, v, 1]^T$ transforms the pixel coordinates into normalized camera coordinates, and $\mathbf{R}$ rotates these coordinates into the world space. The unit-normalized direction $\mathbf{d}'$ is obtained by normalizing $\mathbf{d}$:
\[
\mathbf{d}' = \frac{\mathbf{d}}{\|\mathbf{d}\|}.
\]
The Plücker representation $\mathbf{p}$ is then given by:
\[
\mathbf{s} = (\mathbf{o} \times \mathbf{d}', \mathbf{d}'),
\]
where $\mathbf{o} \times \mathbf{d}'$ represents the moment of the line, calculated as the cross product of the camera origin and the unit direction vector. We generate a per-frame Plücker tensor $\mathbf{P}_i \in \mathbb{R}^{6 \times h \times w}$, ensuring that its spatial dimensions $h$ and $w$ align with those of the video, which is favorable for conditioning with ControlNet.

\section{Projection Formulation for the Mean of 3DGS}

In this section, we describe how the XYZ positions of the 3DGS are obtained through Plücker embedding. Plücker embedding defines the ray origin and direction for each pixel, allowing us to map the network's output depth to spatial coordinates.

The Plücker embedding provides a representation of lines in 3D space using two vectors: the ray origin \(\mathbf{o}\) and the ray direction \(\mathbf{d}\). For each pixel, these vectors define a line in space. The depth value \(z\) output by our network can be used to compute the XYZ position \(\mathbf{p}\) of the 3DGS using the following mapping formula:
\[
\mathbf{p} = \mathbf{o} + z \cdot \mathbf{d}.
\]
Here, \(\mathbf{o}\) is the origin of the ray, \(\mathbf{d}\) is the direction of the ray, and \(z\) is the depth value. This formulation allows us to convert depth information into precise spatial coordinates, effectively reconstructing the 3D geometry of the scene.

By leveraging Plücker embedding, our approach ensures that each pixel's depth is accurately projected into 3D space, facilitating the generation of a pixel-aligned 3DGS representation. However, if the generated video latent does not match the camera condition, the projection may lead to degraded geometry, which further affects the rendering quality.

\end{document}